\documentclass[runningheads]{llncs}
\usepackage{graphicx}
\usepackage{comment}
\usepackage{amsmath,amssymb} 
\usepackage{color, colortbl}
\usepackage{pifont}
\usepackage{makecell}
\usepackage{floatrow}
\usepackage{multirow}
\usepackage{adjustbox}
\usepackage{microtype}
\usepackage{subfigure}
\usepackage{booktabs} 
\usepackage{pbox}
\usepackage{multirow}
\usepackage[table]{xcolor}
\usepackage{float}
\floatstyle{plaintop}
\restylefloat{table}
\usepackage{pdfpages}

\usepackage{xspace}
\makeatletter
\DeclareRobustCommand\onedot{\futurelet\@let@token\@onedot}
\def\@onedot{\ifx\@let@token.\else.\null\fi\xspace}

\def\eg{\emph{e.g}\onedot} 
\def\ie{\emph{i.e}\onedot}

\def\etal{\emph{et al}\onedot}
\makeatother
\usepackage{wrapfig,lipsum,booktabs}

\usepackage{lipsum}
\usepackage[]{algorithm2e}

\definecolor{NVblue}{rgb}{0.07, 0.12, 0.83}
\definecolor{BUred}{rgb}{0.8, 0.0, 0.0}
\usepackage[pagebackref=true,breaklinks=true,colorlinks=true,citecolor=NVblue,linkcolor=BUred,bookmarks=false]{hyperref}

\newfloatcommand{capbtabbox}{table}[\captop][\FBwidth]

\definecolor{mint}{rgb}{0.24, 0.71, 0.54}
\definecolor{orangepeel}{rgb}{1.0, 0.62, 0.0}
\definecolor{awesome}{rgb}{1.0, 0.13, 0.32}

\newcommand{\cmark}{\color{mint}\ding{52}}%
\newcommand{\xmark}{\color{awesome}\ding{55}}%

\newcommand{\rcmark}{\color{awesome}\ding{52}}%
\newcommand{\gxmark}{\color{mint}\ding{55}}%

\usepackage[width=122mm,left=12mm,paperwidth=146mm,height=193mm,top=12mm,paperheight=217mm]{geometry}

\newcommand\blfootnote[1]{%
  \begingroup
  \renewcommand\thefootnote{}\footnote{#1}%
  \addtocounter{footnote}{-1}%
  \endgroup
}

\begin{document}

\pagestyle{headings}
\mainmatter
\def\ECCVSubNumber{1766}  


\title{Unsupervised Cross-Modal Alignment for Multi-Person 3D Pose Estimation}

\titlerunning{Unsupervised Cross-Modal Alignment for Multi-Person 3D Pose Estimation} 
\authorrunning{Kundu \etal} 

\author{Jogendra Nath Kundu* \and Ambareesh Revanur* \and  Govind Vitthal Waghmare  \and \\ Rahul Mysore Venkatesh \and R. Venkatesh Babu}


\institute{Video Analytics Lab, Indian Institute of Science, Bangalore}

\maketitle

\begin{abstract}

We present a deployment friendly, fast bottom-up framework for multi-person 3D human pose estimation. We adopt a novel neural representation of multi-person 3D pose which unifies the position of person instances with their corresponding 3D pose representation. This is realized by learning a generative pose embedding which not only ensures plausible 3D pose predictions, but also eliminates the usual keypoint grouping operation as employed in prior bottom-up approaches. Further, we propose a practical deployment paradigm where paired 2D or 3D pose annotations are unavailable. In the absence of any paired supervision, we leverage a frozen network, as a teacher model, which is trained on an auxiliary task of multi-person 2D pose estimation. We cast the learning as a cross-modal alignment problem and propose training objectives to realize a shared latent space between two diverse modalities. We aim to enhance the model's ability to perform beyond the limiting teacher network by enriching the latent-to-3D pose mapping using artificially synthesized multi-person 3D scene samples. Our approach not only generalizes to in-the-wild images, but also yields a superior trade-off between speed and performance, compared to prior top-down approaches. Our approach also yields state-of-the-art multi-person 3D pose estimation performance among the bottom-up approaches under consistent supervision levels. \blfootnote{{*Equal contribution. $|$ {\textit{Webpage}}: \url{https://sites.google.com/view/multiperson3D}
}} 
\end{abstract}

\section{Introduction}
Multi-person 3D human pose estimation aims to simultaneously isolate individual persons and estimate the location of their semantic body joints in a 3D space. This challenging task can aid a wide range of applications related to human behavior understanding such as surveillance~\cite{Zheng_2017_CVPR}, group activity recognition \cite{luvizon20182d}, sports analytics \cite{ibrahim2016hierarchical}, etc. Existing multi-person pose estimation approaches can be broadly classified into two categories namely, top-down and bottom-up. In top-down approaches \cite{rogez2017lcr,rogez2019lcr,Dabral2019MultiPerson3H,moon2019camera}, the first step is to detect persons using an off-the-shelf detector which is followed by predicting a 3D pose for each person using a single-person 3D pose estimator. Such approaches \cite{rogez2017lcr,rogez2019lcr} are usually incapable of inferring absolute camera-centered distance of each human as they miss the global context. In contrast, the bottom-up approaches \cite{mehta2018single} first locate the body joints, and then assign them to each individual person via a keypoint grouping operation. 
The bottom-up approaches yield suboptimal results as compared to top-down approaches, but have a superior run-time advantage against top-down methods~\cite{kocabas2018multiposenet,redmon2016you}. In this paper, we aim to leverage the computational advantage of bottom-up approaches while effectively eliminating the keypoint grouping operation via an efficient 3D pose representation. This results in a substantial gain in performance while maintaining an optimal computational overhead.

\begin{figure*}[t]
\TopFloatBoxes
\begin{floatrow}

\ffigbox[1.2\linewidth]{
\includegraphics[width=1.0\linewidth]{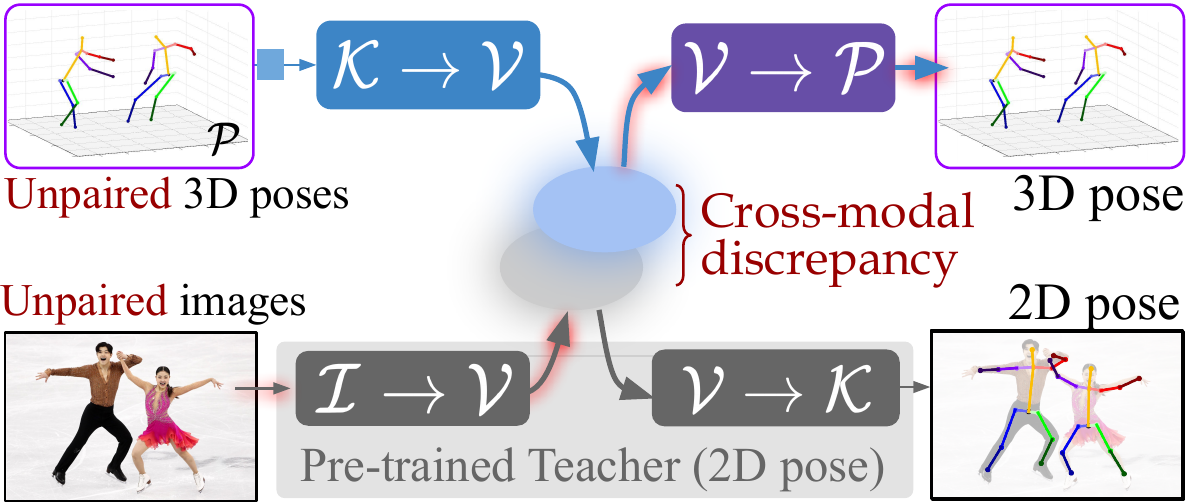}
}{%
  \caption{We aim to realize a shared latent space $\mathcal{V}$ which embeds samples from varied input modalities \ie the unpaired images and unpaired 3D poses. Auto-encoding pathway:  $\mathcal{K}\rightarrow\mathcal{V}\rightarrow\mathcal{P}$. Distillation pathway: from $\mathcal{I}\rightarrow\mathcal{V}\rightarrow\mathcal{K}$ to camera projection of $\mathcal{I}\rightarrow\mathcal{V}\rightarrow\mathcal{P}$. Inference: $\mathcal{I}\rightarrow\mathcal{V}\rightarrow\mathcal{P}$ (red shadow).
  }
  \label{fig:concept_fig}
}

\ffigbox[0.8\linewidth]{
    \includegraphics[width=1.0\linewidth]{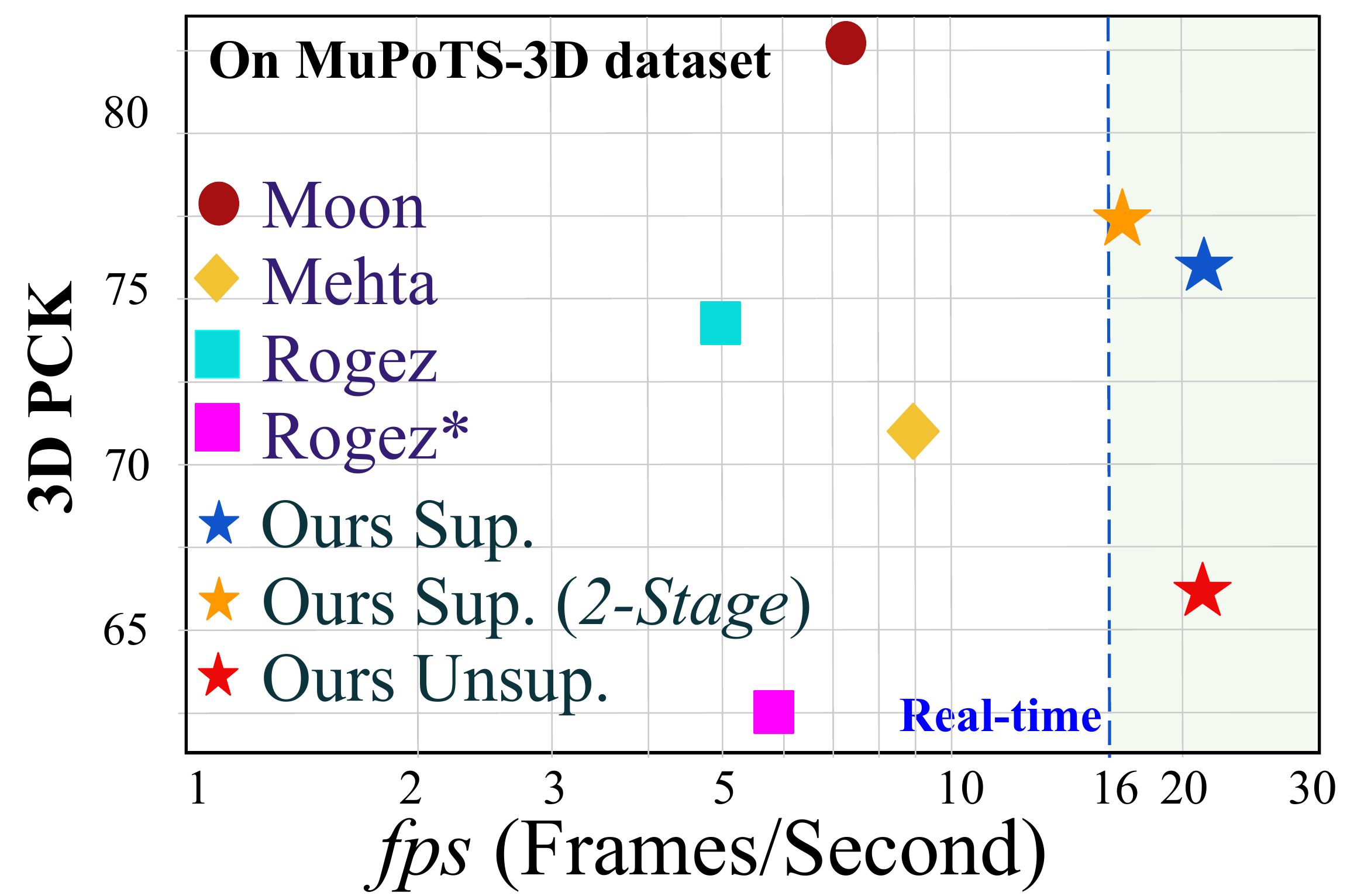}
}{%
    \caption{We achieve a superior trade-off between speed and performance against the prior arts (Rogez\cite{rogez2019lcr}, Rogez*\cite{rogez2017lcr}, Mehta\cite{mehta2018single}, Moon\cite{moon2019camera}). See Section \ref{sec:discussion}
    }
    \label{fig:performance}
}

\end{floatrow}
\end{figure*}

Almost all multi-person 3D pose estimation approaches access large-scale datasets with 3D pose annotations. However, owing to the difficulties involved in capturing 3D pose in wild outdoor environments, many of the 3D pose datasets are captured in indoor settings. This restricts diversity in the corresponding images (\ie limited variations in background, attires and pose performed by actors) \cite{ionescu2013human3,joo2015panoptic}. However, 2D keypoint annotations~\cite{nath2018object,kundu2018ispa} are available even for in-the-wild multi-person outdoor images. Hence, several approaches aim to design 2D-to-3D pose lifters \cite{chen2019unsupervised_cvpr,martinez2017simple} by relying on an off-the-shelf, Image-to-2D pose estimator. Such approaches usually rely on geometric self-consistency of the projected 2D pose obtained from the lifter output, while imposing adversarial prior to assure plausible 3D pose predictions \cite{chen2019unsupervised_cvpr,kanazawa2018end}. 
However, the generalizability of such approaches is limited owing to the dataset bias exhibited by the primary Image-to-2D pose estimator which is trained in a fully-supervised fashion.

\textbf{Our problem setting.} 
Consider a scenario where a pretrained Image-to-2D pose estimator is used for the goal task of 3D pose estimation. There are two challenges that must be tackled. First, a pretrained Image-to-2D estimator would exhibit a dataset bias towards the training data. Thus, the deployment of such a model in an unseen environment (\eg dancers in unusual costumes) is not guaranteed to result in an optimal performance. This curtails the learning of the 2D-to-3D pose lifter, especially in the absence of paired images from the unseen environment. Second, along with the Image-to-2D model, one can not expect to be provided with its labeled training dataset owing to proprietary \cite{nayak2019zero,dfkd} or even memory \cite{lwm,lwf} constraints. Considering these two challenges, the problem boils down to performing domain adaptation~\cite{kundu2019adapt} by leveraging the pretrained Image-to-2D network (\textit{a.k.a} the teacher network) in an unsupervised fashion, \ie \textit{in the absence of any paired 2D or 3D pose annotations}. Further, acknowledging the limitations of existing 2D-to-3D pose lifters, we argue that the 3D pose lifter should access the latent convolutional features instead of the final 2D pose output; owing to its greater task transferability \cite{long2015learning}.

Though it is easy to obtain unpaired multi-person images, acquiring a dataset of unpaired multi-person 3D pose is inconvenient. To this end, we synthesize multi-person 3D scenes by randomly placing the single-person 3D skeletons in a 3D grid as shown in Fig.~\ref{fig:aae_and_syn_scene}{\color{red}B}. We also formalize a systematic way to synthesize single-person 3D pose by accessing plausible ranges of parent-relative joint angle limits provided by biomechanic experts. This eradicates our dependency even on an unpaired 3D skeleton dataset. Our idea of creating artificial samples stems from the concept of domain randomization \cite{peng2018sim,tobin2017domain} which is shown to be effective for generalizing deep models to unseen target environments. The core hypothesis is that the multi-person 3D pose distribution characterized by the artificially synthesized 3D pose scenes would subsume the unknown target distribution. Note that the proposed joint angle sampling would allow sampling of minimal implausible single-person poses as it does not adhere to the strong pose-conditioned joint angle priors formalized by Akhter~\etal~\cite{akhter2015pose}.

We posit the learning framework as a cross-modal alignment problem (see Fig.~\ref{fig:concept_fig}). To this end, we aim to realize a shared latent space $\mathcal{V}$, which embeds samples from varied input modalities~\cite{chung2018unsupervised}, such as unpaired multi-person image $\mathcal{I}$ and unpaired multi-person 2D pose $\mathcal{K}$ (\ie camera projection on multi-person 3D pose $\mathcal{P}$). Our training paradigm employs an auto-encoding loss on $\mathcal{P}$ (via $\mathcal{K}\rightarrow \mathcal{V} \rightarrow \mathcal{P}$ pathway), a distillation loss on $\mathcal{K}$  (via $\mathcal{I}\rightarrow\mathcal{V}\rightarrow \mathcal{P} \rightarrow \mathcal{K}$ pathway) and an additional adaptation loss (non-adversarial) to minimize the cross-modal discrepancy at the latent space $\mathcal{V}$. In further training iterations, we stop the limiting distillation loss and fine-tune the model on a self-supervised criteria based on the equivariance property \cite{schmidt2012learning} of spatial-transformations on the image and its corresponding 2D pose representation. Extensive experiments of our ablations and comparisons against prior arts establish the superiority of this approach. In summary, our contributions are as follows:
\begin{itemize}
    \item We propose an efficient bottom-up architecture that yields fast and accurate single-shot multi-person 3D pose estimation performance with structurally infused articulation constraints to assure valid 3D pose output. In absence of paired supervision we cast the learning as a cross-modal alignment problem and propose training objectives to realize a shared latent space between two diverse data-flow pathways.
    \item We enhance the model's ability to perform even beyond the limiting teacher network as a result of the enriched latent-to-3D-pose mapping using artificially synthesized multi-person 3D scene samples.
    \item Our approach not only yields \textit{state-of-the-art} multi-person 3D pose estimation performance among the prior bottom-up approaches but also demonstrates a superior trade-off between speed and performance.
\end{itemize}

\section{Related Work}

\noindent Multi-person 2D pose estimation works can be broadly classified into top-down and bottom-up methods. Top-down methods such as \cite{chen2018cascaded,newell2016stacked,huang2017coarse,xiao2018simple} first detect the persons in the image and then estimate their poses. On the other hand, bottom-up methods \cite{newell2017associative,pishchulin2016deepcut,insafutdinov2016deepercut,cao2017realtime,nie2019single} predict the pose of all persons in a single-shot. Cao \etal \cite{cao2017realtime} use a non-parametric representation Part Affinity Field (PAF) and Part Confidence Map (PCM) to learn association between 2D keypoints and persons in the image. Similarly, Kocabas \etal \cite{kocabas2018multiposenet} proposed a bottom-up approach using pose residual network for estimating both keypoints and human detections simultaneously. 

\begin{wraptable}[14]{r}{0.45\textwidth}

\caption{{
Characteristic comparison against prior works. \textbf{without paired supervision} implies the method does not need access to annotations.
}}

\centering
\label{tab:characteristic}
\setlength{\tabcolsep}{2.8pt}

\resizebox{0.99\columnwidth}{!}
{
    \begin{tabular}{l|c|c|c}
    
    \hline
    Methods & \makecell{Single \\ shot} & \makecell{\textit{w/o} paired \\ supervision} & \makecell{Camera \\ centric} \\
   
    \hline
    
    Rogez \cite{rogez2017lcr} & \xmark & \xmark  & \xmark\\
    Mehta \cite{mehta2018single}&   \cmark & \xmark  & \xmark\\
    Rogez \cite{rogez2019lcr} &  \xmark & \xmark  & \xmark\\
    Dabral \cite{Dabral2019MultiPerson3H} &  \xmark & \xmark & \cmark \\
    Moon \cite{moon2019camera} &  \xmark & \xmark & \cmark \\
    \hline
    \rowcolor{gray!14}
    Ours  &  \cmark & \cmark  & \cmark \\
    \hline
    \end{tabular}
   
}
\end{wraptable}

Many approaches have been proposed for solving the problem of  single-person 3D human pose estimation \cite{sun2017compositional,kundu2020self,kundu2020unsupervised,kundu2020kinematic,pavlakos2017coarse,yasin2016dual}. Vnect \cite{mehta2017vnect} is the first realtime 3D human pose estimation work that infers the pose by parsing location-maps and joint-wise heatmaps. Martinez \etal \cite{martinez2017simple} proposed an effective approach to directly lift the ground-truth 2D poses to 3D poses. 
Few methods have been proposed so far for Multi-person 3D pose estimation. In  \cite{rogez2017lcr,rogez2019lcr}, Rogez \etal proposed a top-down approach based on localization, classification and regression of 3D joints. These modules are pipelined to predict the final pose of all persons in the image. Mehta \etal \cite{mehta2018single} proposed a single-shot approach to infer 3D poses of all people in the image using PAF-PCM representation. To handle occlusions, they introduced Occlusion Robust Pose Maps (ORPM) which allows full body pose inference under occlusions. Moon \etal \cite{moon2019camera} proposed the first top-down camera-centered 3D pose estimation. Their framework contains three modules: DetectNet localizes multiple persons in the image, RootNet estimates camera-centered depth of root joint and PoseNet estimates root relative 3D pose of the cropped person. In RootNet, they use pinhole camera projection model to estimate absolute camera-centered depth. Dabral \etal \cite{Dabral2019MultiPerson3H} proposed a 2D to 3D lifting based approach for camera-centric predictions.  
Rogez \etal \cite{rogez2017lcr,rogez2019lcr} and Moon \etal \cite{moon2019camera} crop the detected person instances from the image and they do not leverage the global context information. 
All prior state-of-the-art works \cite{rogez2017lcr,rogez2019lcr,mehta2018single,Dabral2019MultiPerson3H,moon2019camera} require paired supervision. See Table \ref{tab:characteristic} for a characteristic comparison against prior works.

\noindent \textbf{Cross-modal distillation.} Gupta \etal \cite{gupta2016cross} proposed a novel method for enabling cross-modal transfer of supervision for tasks such as depth estimation. They propose alignment of representations from a large labeled modality to a sparsely labeled modality. In \cite{spurr2018cross}, Spurr \etal demonstrated the effectiveness of cross-modal alignment of latent space for the task of hand pose estimation. In a related work \cite{pilzer2019refine}, Pilzer \etal proposed an unsupervised distillation based depth estimation approach via refinement of cycle-inconsistency.

\section{Approaches}

Our prime objective is to realize a learning framework for the task of multi-person 3D pose estimation without accessing any paired data (\ie images with the corresponding 2D or 3D pose annotations). To achieve this, we plan to distill the knowledge from a frozen teacher network which is trained for an auxiliary task of multi-person 2D landmark estimation. Furthermore, in contrast to the general top-down approaches in fully-supervised scenarios, we propose an effective single-shot, bottom-up approach for multi-person 3D pose estimation. Such an architecture not only helps us maintain an {optimal computational overhead} but also lays a suitable ground for cross-modal distillation.

\subsection{Architecture}
Aiming to design a single-shot end-to-end trainable architecture, we draw motivation from the real-time object detectors such as YOLO \cite{redmon2016you}. The output layer in YOLO divides the output spatial map into a regular grid of cells. The multi-dimensional vector at each grid location broadly represents two important attributes. Firstly, a confidence value indicating the existence of an object centroid in the corresponding input image patch upon registering the grid onto the spatial image plane. Secondly, a parameterization of the object properties, such as class probabilities and attributes related to the corresponding bounding box. In similar lines, for multi-person 3D pose estimation, each grid location of the output layer represents a heatmap indicating existence of a human pelvis location (or root) followed by a \textit{parameterization} of the corresponding root-relative 3D pose. Here, the major challenge is how to parameterize root-relative human 3D pose in the efficient manner. We explicitly address it in the following subsection.

\subsubsection{Parameterizing 3D pose via pose embedding.}
\label{sec:aae}
\begin{figure}[t]
    \centering
    \includegraphics[width=0.98\textwidth]{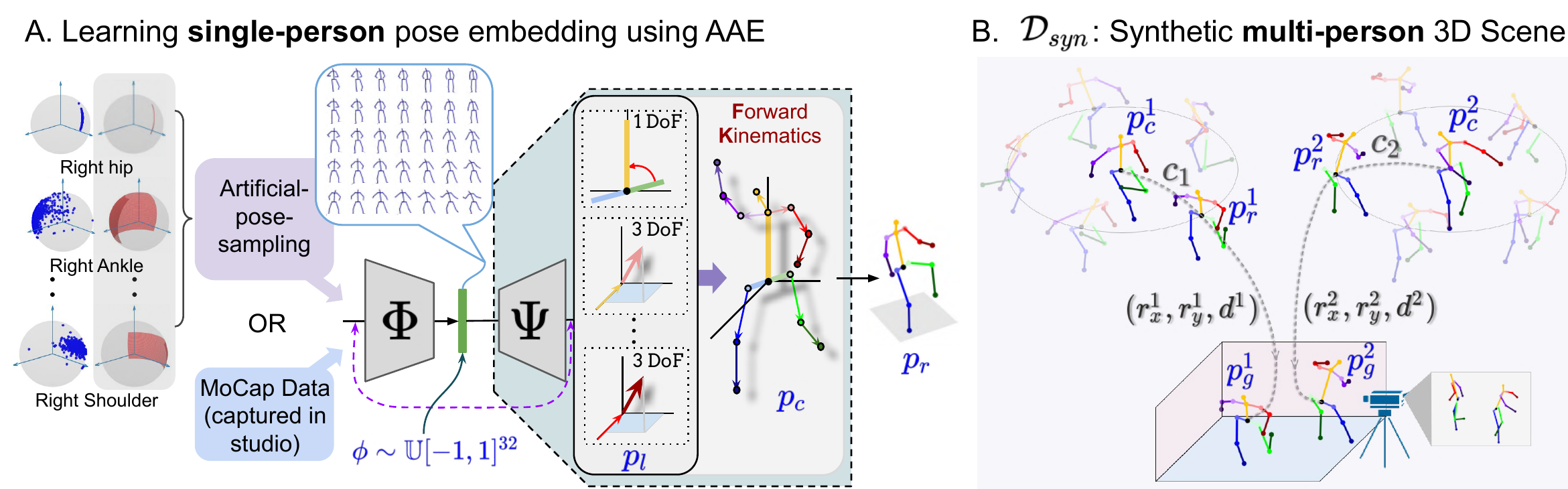}
   
    \caption{\textbf{A}. Learning continuous pose-embedding on MoCap or Artificially sampled pose dataset. 
    \textbf{B}. Creating $\mathcal{D}_{syn}$: Each canonical pose  $p_c$ is rigidly transformed through rotation and translation operation to form random 3D scenes. 
    }
    \label{fig:aae_and_syn_scene}
\end{figure}

Root relative human 3D pose follows a complex structured articulation. Moreover, defining a parameterization procedure without accounting for the structural plausibility of the 3D pose would further add up to the inherent 2D to 3D ambiguity. Acknowledging this, we aim to devise a parameterization which selectively decodes anthropomorphically plausible human poses spanning a continuous latent manifold (see Fig. \ref{fig:aae_and_syn_scene}{\color{red}A}). One of the effective ways to realize the above objective is to train a generative network~\cite{kundu2019gan} which models the most fundamental form of human pose variations. Thus, we disentangle the root-relative pose $p_r$ into its rigid and non-rigid factors. The non-rigid factor, also known as the canonical pose $p_c$ is designed to be view-invariant. The rigid transformation is defined by the parameters $c$ as required for the corresponding rotation matrix. In further granularity, according to the concept of forward kinematics~\cite{zhou2016deep}, movement of each limb is constrained by the parent-relative joint-angle limits and the scale invariant fixed relative bone lengths. Thus, the unit vectors corresponding to each joint defined at their respective parent-relative local coordinate system~\cite{akhter2015pose} is regarded as the most fundamental form of 3D human pose which is denoted by $p_l$. Note that, the transformation $p_l\rightarrow p_c$ is a fully-differentiable series of forward kinematic operations. We train a generative network~\cite{kundu2019bihmp,kundu2019unsupervised} following the learning procedure of adversarial auto-encoder $\{\Phi,\Psi\}$ (AAE~\cite{makhzani2015adversarial})  on samples of $p_l$ acquired from either a MoCap \cite{cmumocap} dataset or via a proposed \textit{Artificial-pose-sampling} procedure (see Fig.~\ref{fig:aae_and_syn_scene}{\color{red}A}). 
We consider a uniform prior distribution \ie $\mathbb{U}[-1,1]^{32}$. This ensures that any random vector $\phi\in[-1,1]^{32}$ decodes (via $\Psi$) an anthropomorphically plausible human pose. (See Suppl)

In the proposed \textit{Artificial-pose-sampling} procedure, we use a set of joint angle limits (4 angles \ie the allowed range of polar and azimuthal angles in the parent relative local pose representation) provided by the biomechanic experts. The angle for each limb is independently sampled from a uniform distribution defined by the above range values (see the highlighted regions on the sphere for each body joint in Fig.~\ref{fig:aae_and_syn_scene}{\color{red}A}). Note that, the proposed joint angle sampling would allow sampling of minimal implausible single-person poses as it does not adhere to the pose-conditioned joint angle limits formalized by Akther~\etal~\cite{akhter2015pose}. (See Suppl)

\subsubsection{Neural representation of multi-person 3D pose.}
\begin{figure*}[t]
    \centering
    \includegraphics[width=0.98\textwidth]{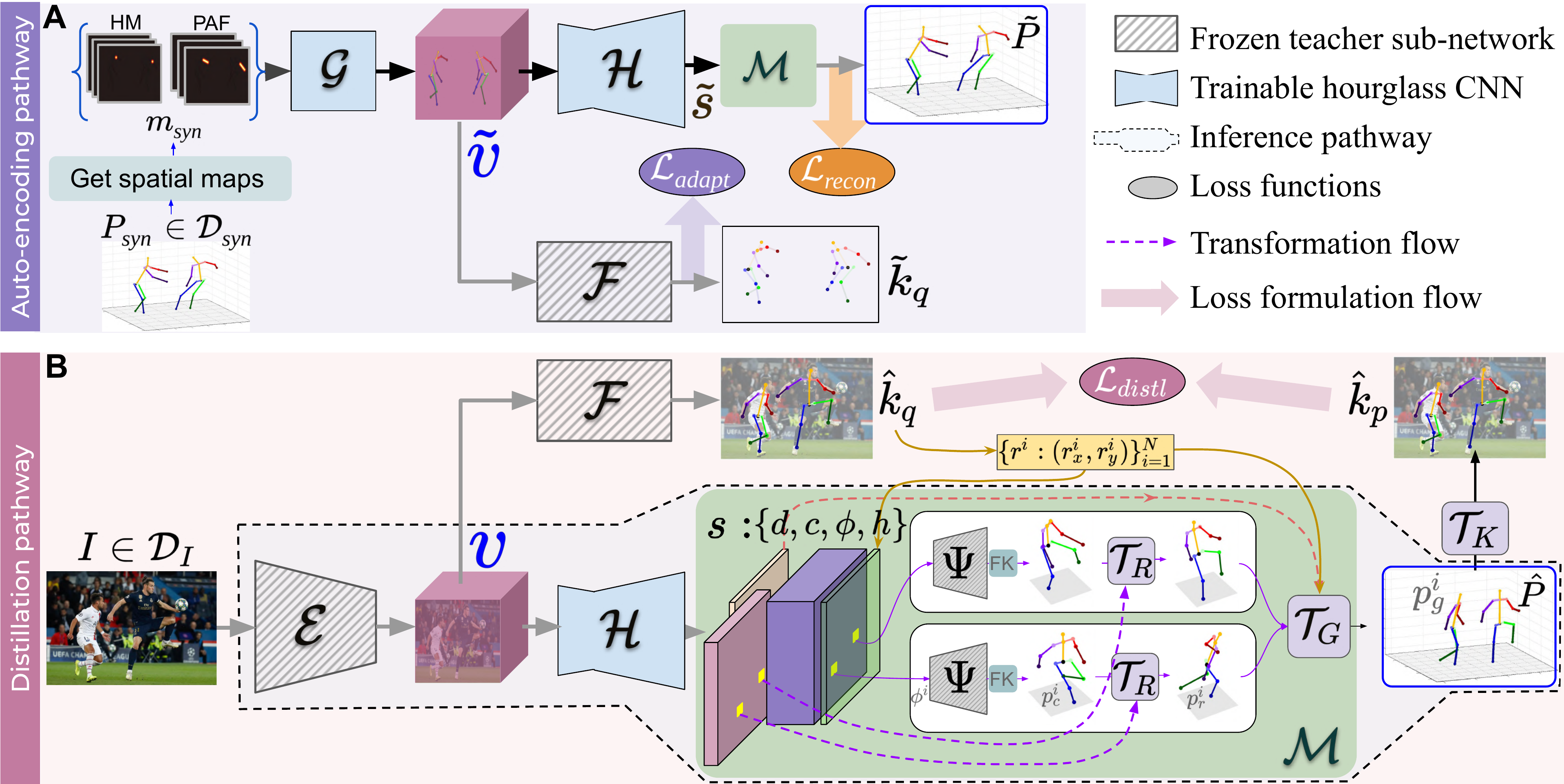}
    \caption{Proposed data-flow pathways.
    Distillation is performed from the teacher, $\{\mathcal{E}, \mathcal{F}\}$ to the student $\{\mathcal{H}\}$. Weights of 
    $\mathcal{H}$ and $\mathcal{F}$ are shared across both the pathways.
    }
    \label{fig:architecture}
\end{figure*}

The last layer output of the single-shot latent to multi-person 3D pose mapper $\mathcal{H}$, denoted as $s$, is a 3D tensor of size $H\times W\times M$ (see block $\mathcal{M}$ Fig.~\ref{fig:architecture}{\color{red}B}). The number of channels constitutes of 4 distinct components. The $M$ dimensional vector for each grid location $r^i$ constitutes of 4 distinct components viz, \textbf{a)} a scalar heatmap intensity indicating existence of a skeleton pelvis denoted as $h^i$, \textbf{b)} a 32 dimensional 3D pose embedding $\phi^i$, \textbf{c)} 6 dimensional rigid transformation parameters $c^i$ ($\sin$ and $\cos$ component of 3 rotation angles), and \textbf{d)} a scalar absolute depth $d^i$ associated with the skeleton pelvis. Note that, the last 3 components are interpretable only in presence of a pelvis at the corresponding grid location as denoted by the first component. Here, $\phi^i$ is obtained through a \textit{tanh} nonlinearity thus constraining it to decode (via frozen $\Psi$ AAE from Section \ref{sec:aae})  only plausible 3D human poses.

The model accesses a set of 2D pelvis key-point locations belonging to each person in the corresponding input image, denoted as $\{r^i\}_{i=1}^{N}$. Here, $N$ denotes the total number of persons. These spatial locations are obtained either as estimated by the teacher network or from the ground-truth depending on its availability. For each selected location $r^i$, the corresponding $\phi^i$ and $c^i$ are pooled from the relevant grid location to decode (via $\Psi$) the corresponding root-relative 3D pose, $p_r^i$. First, the canonical pose, $p_c^i$ is obtained by applying forward kinematics (denoted as FK in Fig. \ref{fig:architecture}{\color{red}B} in module $\mathcal{M}$) on the decoded local vectors obtained from the pose embedding $\phi^i$. Following this, $p_r^i$ is obtained after performing rigid transformation using $c^i$, \ie $\mathcal{T}_R$ in Fig.~\ref{fig:architecture}{\color{red}B}. Finally, the global 3D pose scene, $\hat{P} = \{p_g^i\}_{i=1}^N$, is constructed by translating the root-relative 3D poses to their respective root locations in the camera centered global coordinate system, \ie $\mathcal{T}_G$ in Fig.~\ref{fig:architecture}{\color{red}B}. The 3D translation for each person $i$ is obtained using $(r_x^i, r_y^i, d^i)$, {where $r_x^i$ and $r_y^i$ are the X and Y component obtained as a transformation of the spatial root location $r^i$}. In Fig.~\ref{fig:architecture}{\color{red}B}, the series of fixed (non-trainable) differentiable operations to obtain $\hat{P}$ from the CNN output $s$ is denoted as $\mathcal{M}$. A {weak perspective camera transformation}, $\mathcal{T}_K$, of $P$ provides us the corresponding multi-person 2D key-points denoted by $\hat{k}_p$.

\noindent
\textbf{Inference.} During inference, $(r_x^i, r_y^i)$ is obtained from the heatmap channel $h$ predicted at the output of $\mathcal{F}$. We follow the non-maximum suppression algorithm {inline with} Cao~\etal~\cite{cao2017realtime} to obtain a set of spatial root locations belonging to each person. Thus, the inference pathway during testing is as follows, $\hat{P} = \mathcal{M}\circ \mathcal{H} \circ \mathcal{E}(I)$.

\subsection{Learning cross-modal latent space}

We posit the learning framework as a {cross-modal alignment} problem. Moreover, we aim to realize a shared latent space, $\mathcal{V}$ which embed samples from varied modality spaces, such as multi-person image $\mathcal{I}$, multi-person 2D pose $\mathcal{K}$, and multi-person 3D pose $\mathcal{P}$. However, in absence of labeled samples (or paired samples) an intermediate representation of the frozen teacher network is treated as the shared latent embedding. Following this, separate mapping networks are trained to encode or decode the latent representation to various source modalities. Note that, the teacher network already includes the mapping of image to the latent space, $\mathcal{E}:\mathcal{I}\rightarrow\mathcal{V}$ and latent space to multi-person 2D pose, $\mathcal{F}:\mathcal{V}\rightarrow\mathcal{K}$. We train two additional mapping networks, viz. a) multi-person 2D pose to latent space, $\mathcal{G}:\mathcal{K}\rightarrow{\mathcal{V}}$ and b) latent space to multi-person 3D pose, $(\mathcal{M}\circ\mathcal{H}): {\mathcal{V}}\rightarrow\mathcal{P}$. Also note that, $(\mathcal{T}_K\circ\mathcal{M}\circ\mathcal{H}): {\mathcal{V}}\rightarrow\mathcal{K}$.

\textbf{Available Datasets.}  
We have access to two unpaired datasets viz. a) unpaired multi-person images $I\sim\mathcal{D}_I$ and b) unpaired multi-person 3D pose samples $P_{syn}\sim\mathcal{D}_{syn}$. Though it is easy to get hold of unpaired multi-person images, acquiring a dataset of unpaired multi-person 3D pose is inconvenient. Acknowledging this, we propose a systematic procedure to synthesize a large-scale multi-person 3D pose dataset from a set of plausible single-person 3D poses. A multi-person 3D pose sample constitute of a certain number of persons (samples of $p^i_l$) with random rigid transformations ($c^i$) placed at different locations (\ie $r_x^i, r_y^i, d^i$) 
in a 3D room. This is illustrated in Fig.~\ref{fig:aae_and_syn_scene}{\color{red}B}. Here, samples of $p_l$ can be obtained either from a MoCap dataset or by following \textit{Artificial-pose-sampling}.

Broadly, we use two different data-flow pathways as shown in Fig.~\ref{fig:architecture}. Here, we discuss how these pathways support an effective cross-modal alignment.

\textbf{a) Cross-modal distillation pathway for $I\sim\mathcal{D}_{I}$.} The objective of distillation pathway is to instill the knowledge of mapping an input RGB image to the corresponding multi-person 2D pose (\ie from the teacher network $\hat{k}_q = \mathcal{F}(v)$ where $v=\mathcal{E}(I)$) into the newly introduced 3D pose estimation pipeline. Here, $\hat{k}_q$ is obtained after performing bipartite matching inline with Cao~\etal~\cite{cao2017realtime}. We update the parameters of $\mathcal{H}$ by imposing a distillation loss between $\hat{k}_q$ and the perceptively projected 2D pose $\hat{k}_p=\mathcal{T}_K\circ\mathcal{M}\circ\mathcal{H}(v)$, \ie $\boldsymbol{\mathcal{L}_{\textit{distl}}=\vert \hat{k}_p - \hat{k}_q \vert}$.

\textbf{b) Auto-encoding pathway for $P_\textit{syn}\sim\mathcal{D}_{\textit{syn}}$.} In the auto-encoding pathway, the objective is to reconstruct back the synthesized samples of multi-person 3D poses via the shared latent space. Owing to the spatially structured latent representation, for each non-spatial $P_\textit{syn}$ we first generate the corresponding multi-person spatial heatmap (HM) and Part Affinity Map (PAF) inline with Cao~\etal~\cite{cao2017realtime}, denoted by $m_\textit{syn}$ in Fig.~\ref{fig:architecture}{\color{red}A}. Note that $m_{syn}$ represents the 2D keypoint locations of $k_{syn}$ which is the obtained as the camera projection of the $P_{syn}$. Following this, we obtain $\tilde{P}=\mathcal{M}\circ\mathcal{H}(\tilde{v})$ where $\tilde{v}=\mathcal{G}(m_\textit{syn})$. Parameters of both $\mathcal{G}$ and $\mathcal{H}$ are updated to minimize $\boldsymbol{\mathcal{L}_{\textit{recon}}=\vert P_\textit{syn} - \tilde{P} \vert}$.

\textbf{c) Cross-modal adaptation.} 
Notice that, $\mathcal{H}$ is the only common model updated in both pathways. Here, $\mathcal{L}_{\textit{distl}}$ is computed against the noisy teacher prediction that too in the 2D pose space. In contrast, $\mathcal{L}_{\textit{recon}}$ is computed against the true ground-truth 3D pose thus devoid of the inherent 2D to 3D ambiguity. As a result of this disparity, the model $\mathcal{H}$ differentiates between the corresponding input distributions, \ie between $\mathbb{P}(v)$ and $\mathbb{P}(\tilde{v})$, thereby learning separate strategies favouring the corresponding learning objectives. To minimize this discrepancy, we rely on the frozen teacher sub-network $\mathcal{F}$. We hypothesize that, the energy computed via $\mathcal{F
}$, \ie $\vert \mathcal{F}(\tilde{v}) - m_\textit{syn} \vert$ would be low if the associated input distribution of $\mathcal{F}$, \ie $\mathbb{P}(v=\mathcal{E}(I))$ aligns with the output distribution of $\mathcal{G}$, \ie $\mathbb{P}(\tilde{v}=\mathcal{G}(m_\textit{syn}))$. Accordingly, we propose to minimize $\boldsymbol{\mathcal{L}_\textit{adapt}=\vert \mathcal{F}\circ\mathcal{G}
(m_\textit{syn}) - m_\textit{syn} \vert}$ to realize an effective cross-modal alignment.

\noindent
\textbf{Training phase-1} We update $\mathcal{G}$ and $\mathcal{H}$ to minimize all the three losses discussed above, \ie $\mathcal{L}_\textit{recon}$, $\mathcal{L}_\textit{distl}$ and $\mathcal{L}_\textit{adapt}$ each with different Adam~\cite{kingma2014adam} optimizers.

\subsection{Learning beyond the teacher network}
We see a clear limitation in the learning paradigm discussed above. The inference performance of the final model is limited by the dataset bias infused in the teacher network. We recognize $\mathcal{L}_\textit{distl}$ as the prime culprit which limits the ability of $\mathcal{H}$ by not allowing it to surpass the teacher's performance. Though one can rely on $\mathcal{L}_\textit{recon}$ to further improve $\mathcal{H}$, this would degrade performance in the inference pathway as a result of increase in discrepancy between $v$ and $\tilde{v}$. Considering this, we propose to freeze $\mathcal{G}$ thereby freezing its output distribution $\mathbb{P}(\tilde{v}=\mathcal{G}(m_\textit{syn}))$ in the second training phase. 

Furthermore, in absence of the regularizing $\mathcal{L}_\textit{distl}$ we use a self-supervised consistency loss to regularize $\mathcal{H}$ for the unpaired image samples. For each image $I$ we form a pair $(I,I^\prime)$ where $I^\prime =T_s(I)$ is the spatially transformed version (\ie image-flip, random-crop, or in-place rotation) of $I$. Here, $T_s$ represents the differentiable spatial transformation. Next, we propose a consistency loss based on the equivariance property~\cite{schmidt2012learning} of the corresponding multi-person 2D pose, \ie
$$
\mathcal{L}_\textit{ss}=\vert T_s\circ\mathcal{T}_K\circ\mathcal{M}\circ\mathcal{H}\circ\mathcal{E}(I) - \mathcal{T}_K\circ\mathcal{M}\circ\mathcal{H}\circ\mathcal{E}\circ T_s(I)  \vert
$$
The above loss is computed at the root-locations extracted using the teacher network for the original image $I$. Whereas, for $I^\prime$ we use the spatial transformation $T_s$ on the extracted root locations of the original image.  

\noindent
\textbf{Training phase-2} We update the parameters of $\mathcal{H}$ ($\mathcal{G}$ is kept frozen from the previous training phase) to minimize two loss terms \ie $\mathcal{L}_\textit{recon}$ and $\mathcal{L}_\textit{ss}$. 

\section{Experiments}
In this section, we describe the experiments and results of the proposed approach on several benchmark datasets. Through quantitative and qualitative analysis, we demonstrate the practicality and performance of our method.

\subsection{Implementation Details}
First, we explain the implementation details of synthetic dataset creation. Next, we provide the training details for learning the neural representation. 

\textbf{3D skeleton dataset.} \textit{Artificial-pose-sampling} is performed by sampling uniformly from joint wise angle limits defined at local parent relative \cite{akhter2015pose} spherical coordinate system (see Fig.~\ref{fig:aae_and_syn_scene}{\color{red}A}) \ie $[\theta_1, \theta_2],$ and $[\gamma_1, \gamma_2]$ . For example, right-hip joint $\theta_1=\theta_2=\pi$ (\ie 1-DoF) and $\gamma_1=\pi/3$, $\gamma_2=2\pi/3$ (See Suppl). Using these predefined limits, we construct a full 3D pose (via FK). A total of 1M poses are sampled for training  $\Psi_{arti}$. Further, 100k synthetic multi-person pose scenes are created by sampling upto 4 single-person 3D poses per scene. Note that, the $\mathcal{D}_{syn}$ dataset can also utilize 3D poses from single-person 3D dataset such as Human 3.6M \cite{ionescu2013human3} and MPI-INF-3DHP \cite{mehta2017monocular}, when accessible.

\textbf{Training.} First, we train a pose decoder (see Section \ref{sec:aae}) either on artificial pose dataset ($\Psi_{arti}$) or MoCap 3D dataset ($\Psi_{mocap}$). The AAE modules are trained using a batch size of 32, with a learning rate of 1e-4 using Adam optimizers till convergence (See Suppl). The decoder $\Psi$ is frozen for rest of the training. For training the neural representation, we choose the pretrained network of Cao \etal \cite{cao2017realtime} as the teacher network. We consider upto stage-1 ``conv5-4-CPM" layer of \cite{cao2017realtime} as $\mathcal{E}$. We concatenate the predictions of both heatmap and Part Affinity Field branches to obtain an embedding space of size 28$\times$28$\times$1024. We consider module $ \mathcal{F}$ as from stage-1 ``conv5-5-CPM" layer upto stage-2 ``Mconv7-stage2" layer of \cite{cao2017realtime}. 
Using this teacher model, we train the modules $\{\mathcal{H}, \mathcal{G}\}$ by minimizing the losses $\mathcal{L}_{distl}$, $\mathcal{L}_{recon} $, $ \mathcal{L}_{adapt}$, $ \mathcal{L}_{ss}$ using separate Adam optimizers for each of the losses. We use a learning rate of 1e-4 upto 100k iterations and 1e-5 for the following 500k iterations while using a fixed batch size of 8 throughout the training. Further, we use batches of images from $\mathcal{D}_{syn}$ and  $\mathcal{D}_{I}$ in alternate iterations while training the network. The input image size for $\mathcal{D}_{I}$ is $224\times224\times3$ and input PAF representation \cite{cao2017realtime} is of shape $28\times28\times43$ for $\mathcal{D}_{syn}$. All transformations $\mathcal{T}_K$, $\mathcal{T}_R, \mathcal{T}_G$ have been implemented using TensorFlow and are designed to be completely differentiable end-to-end. We have trained the entire pipeline on a Tesla-V100 GPU card in Nvidia-DGX station (See Suppl).

\begin{table*}[t]

    \setlength{\tabcolsep}{1pt}
    \resizebox{\textwidth}{!}{
    
    \caption{Quantitative analysis of different ablations of our approach on MuPoTS-3D. \textit{Unpaired} means that there is no ground truth annotation available for an image. \textit{Paired} means that there is a corresponding annotation available for an image. 3DPCK is Percentage of Correct 3D Keypoints predicted within 15cm. (higher 3DPCK is better). ``sup." stands for supervision. MuCo-3DHP \cite{mehta2018single} is used in fifth column. Red color indicates that configuration is less preferable for low data regime. (\textit{Best viewed in color}).
    }
    \label{tab:ablation}
    \begin{tabular}{l|c|c|c|c|c}
    \hline
         \multirow{2}{*}{Methods} &  
         \multirow{2}{*}{\makecell{Artificial\\ Poses ($\Psi_{arti}$)}} & \multirow{2}{*}{\makecell{MoCap\\ Poses ($\Psi_{mocap}$)}} &
         \multirow{2}{*}{\makecell{Paired \textbf{multi} \\ \textbf{person} 2D sup.}} &  \multirow{2}{*}{\makecell{Composed \textbf{multi}\\\textbf{person} 3D sup. }} &  \makecell{\multirow{2}{*}{3DPCK}} \\
         &&&&&\\
         \hline \hline
         \multicolumn{6}{c}{\textbf{Ours: Learning without any paired supervision.} Using 2D predictions from teacher}
         
         \\
         \hline
          $\mathcal{L}_{distl}$ (no $\mathcal{D}_{syn}$) & \cmark & \gxmark & \gxmark & \gxmark & 53.3   \\
          ~~~ $+ \mathcal{L}_{recon}$  & \cmark & \gxmark & \gxmark & \gxmark & 57.6    \\
          ~~~~~ $+\mathcal{L}_{adapt}$ & \cmark & \gxmark & \gxmark & \gxmark & 61.9   \\
          ~~~~~~~ $+\mathcal{L}_{ss}$  & \cmark & \gxmark & \gxmark & \gxmark & 64.2   \\
          \textit{Ours-Us}  & \xmark & \rcmark & \gxmark & \gxmark & 66.1   \\
         
         \hline
         \multicolumn{6}{c}{~~~~~~~~\textbf{Ours: Weakly Supervised Learning Methods.} Using paired 2D supervision only ~~~~~~~~} \\
         \hline
          with $\Psi_{arti}$  & \cmark & \gxmark & \rcmark & \gxmark & 66.4  \\
          \textit{Ours-Ws}  & \xmark & \rcmark & \rcmark & \gxmark & 67.9 \\
         \hline
         \hline
         \multicolumn{6}{c}{\textbf{Ours: Supervised Learning Methods.} Using both paired 2D and 3D supervision} \\
         \hline
        No $\mathcal{D}_{syn}$  & \xmark & \rcmark & \rcmark & \rcmark & 71.1 \\
          \textit{Ours-Fs} & \xmark & \rcmark & \rcmark & \rcmark & 75.8  \\
          
    \hline
    \end{tabular}
    }
\end{table*}

\subsection{Ablation Studies}
\label{sec:ablation_studies}
In order to study the effectiveness of our method, we perform extensive ablation study by varying levels of supervision, as shown in Table \ref{tab:ablation}. For all the ablations, we have used MuCo-3DHP images \cite{mehta2018single} as $\mathcal{I}$. Depending on the supervision setting, we either access none (for unsup. setting), a small fraction (semi sup. setting) or a complete set (full sup. setting) of 3D annotations in MuCo-3DHP dataset. 

\textbf{\textit{Ours-Us}} (Using \textit{Unpaired} images only): Our baseline model (see Table \ref{tab:ablation}) trained without accessing any annotated labels gives an overall 3DPCK of 53.3.
We observe that $\mathcal{L}_{adapt}+\mathcal{L}_{ss}$ gives a non-trivial boost of 4-6$\%$. This demonstrates the importance of cross-modal alignment and self-supervised consistency. 

\textbf{\textit{Ours-Ws}} (\textit{Weakly supervised}): When supervised weakly by 2D ground truth ($\mathcal{L}_{2D}=|k_p - \hat{k}_p|$), our approach obtains a 3DPCK of 67.9. Further, the performance of our approach that uses $\Psi_{arti}$ is on par with our performance with $\Psi_{mocap}$ indicating that $\phi_{arti}$ has rich representation space, equivalent to $\phi_{mocap}$. 

\textbf{\textit{Ours-Fs}} (\textit{Fully supervised}):  When we access the full training dataset of MuCo-3DHP and impose a 3D reconstruction loss by using $\mathcal{L}_{3D}=|P-\hat{P}|$, we obtain a 3DPCK of 75.8, which is significantly better than the prior arts.

\begin{figure*}[t]

\begin{floatrow}
\capbtabbox{%
    \resizebox{\textwidth}{!}{
    \caption{Comparison of 3DPCK$_{rel}$ on MuPoTS-3D sequences. Our methods are highlighted in gray background color. Underlined values indicate that our unpaired learning (\textit{Ours-Us}) approach performs better on that sequence. \textit{Ours-Fs} (fully-supervised) achieves state-of-the-art in bottom up methods. \textit{Ours-Us} approach performs competitively even when compared with prior fully supervised approaches. 
    }
  
    \begin{tabular}{l|c|c|c|c|c|c|c|c|c|c|c|c|c|c|c|c|c|c|c|c|c}
         \hline
         Methods&S1&S2&S3&S4&S5&S6&S7&S8&S9&S10&S11&S12&S13&S14&S15&S16&S17&S18&S19&S20&Avg \\
         \hline \hline
         \multicolumn{22}{c}{\textit{\textbf{Accuracy for all groundtruths}}} \\
       \hline
        
        Rogez\cite{rogez2017lcr} &\underline{67.7}&\underline{49.8}&\underline{53.4}&\underline{59.1}&\underline{67.5}&22.8&\underline{43.7}&\underline{49.9}&\underline{31.1}&\underline{78.1}&\underline{50.2}&\underline{51.0}&\underline{51.6}&\underline{49.3}&\underline{56.2}&\underline{66.5}&\underline{65.2}&\underline{62.9}&\underline{66.1}&\underline{59.1}&\underline{53.8}
        \\
        Rogez\cite{rogez2019lcr} & \textbf{87.3}&61.9&67.9&74.6&\textbf{78.8}&48.9&\underline{58.3}&\underline{59.7}&\textbf{78.1}&\textbf{89.5}&69.2&73.8&\textbf{66.2}&56.0&\textbf{74.1}&82.1&78.1&72.6&73.1&61.0&70.6
        \\
        Dabral\cite{Dabral2019MultiPerson3H} &85.1&67.9&\textbf{73.5}&\textbf{76.2}&74.9&52.5&\underline{65.7}&\underline{63.6}&\underline{56.3}&\underline{77.8}&76.4&70.1&65.3&\underline{51.7}&69.5&\textbf{87.0}&82.1&\textbf{80.3}&\textbf{78.5}&\textbf{70.7}&71.3
        \\
        
        Mehta\cite{mehta2018single} &81.0&\underline{60.9}&64.4&63.0&69.1&30.3&\underline{65.0}&\underline{59.6}&64.1&83.9&68.0&68.6&62.3&59.2&70.1&80.0&79.6&\underline{67.3}&\underline{66.6}&67.2&66.0
        \\
    \rowcolor{gray!14} 
        \textit{Ours-Us} &76.8&61.8&61.2&63.0&68.7&20.3&67.3&65.2&59.5&83.6&62.4&66.0&52.7&54.9&57.5&73.6&70.9&70.1&70.4&60.8&63.3\\
    \rowcolor{gray!14}
        \textit{Ours-Ws}& 79.6&62.3&54.2&55.9&69.3&36.1&69.1&67.7&58.4&80.2&75.3&68.7&53.6&56.5&59.6&77.4&76.7&69.6&69.2&64.1&65.2 \\
    \rowcolor{gray!14}
        \textit{Ours-Fs}& 85.5&\textbf{84.1}&66.7&70.5&77.4&\textbf{68.6}&\textbf{74.8}&\textbf{77.9}&69.1&80.0&\textbf{78.4}&\textbf{75.4}&61.1&\textbf{60.9}&71.3&81.4&\textbf{85.1}&73.4&74.9&63.5&\textbf{74.0}\\
        \hline
        
        \multicolumn{22}{c}{\textit{\textbf{Accuracy only for matched groundtruths}}} \\
        \hline
        Rogez\cite{rogez2017lcr} &\underline{69.1}&67.3&\underline{54.6}&\underline{61.7}&74.5&25.2&\underline{48.4}&\underline{63.3}&69.0&\underline{78.1}&\underline{53.8}&\underline{52.2}&60.5&\underline{60.9}&\underline{59.1}&\underline{70.5}&76.0&\underline{70.0}&77.1&81.4&\underline{62.4}
        \\
        Rogez\cite{rogez2019lcr} & \textbf{88.0}&73.3&\textbf{67.9}&\textbf{74.6}&\textbf{81.8}&50.1&\underline{60.6}&\underline{60.8}&\textbf{78.2}&\textbf{89.5}&70.8&74.4&\textbf{72.8}&64.5&\textbf{74.2}&84.9&85.2&78.4&75.8&\underline{74.4}&74.0
        \\        
        Dabral\cite{Dabral2019MultiPerson3H} &85.8&73.6&\underline{61.1}&\underline{55.7}&77.9&53.3&\textbf{75.1}&\underline{65.5}&\underline{54.2}&\underline{81.3}&\textbf{82.2}&71.0&70.1&67.7&69.9&\textbf{90.5}&\textbf{85.7}&\textbf{86.3}&\textbf{85.0}&\textbf{91.4}&74.2
        \\
      
        Mehta\cite{mehta2018single} &81.0&\underline{65.3}&64.6&63.9&75.0&30.3&\underline{65.1}&\underline{61.1}&64.1&83.9&72.4&69.9&71.0&72.9&71.3&83.6&79.6&73.5&78.9&90.9&70.8 \\
        \rowcolor{gray!14}
        \textit{Ours-Us} &76.8&66.6&62.1&63.9&73.5&20.3&67.3&67.8&59.5&83.6&62.4&66.0&56.0&63.5&59.5&75.2&70.9&73.0&73.1&80.8&66.1\\
        \rowcolor{gray!14}
        \textit{Ours-Ws} &79.6&66.0&55.5&56.4&74.8&36.1&69.1&69.6&58.4&80.2&75.3&68.7&56.7&66.4&61.6&78.9&76.7&72.8&71.7&83.0&67.9 \\
        \rowcolor{gray!14}
        \textit{Ours-Fs}& 85.5&\textbf{86.5}&66.7&70.5&81.2&\textbf{68.6}&74.8&\textbf{79.5}&69.1&80.0&78.4&\textbf{75.4}&64.0&\textbf{68.6}&73.7&82.9&85.1&76.4&77.4&72.8&\textbf{75.8}\\
        \hline

    \end{tabular}
      \label{tab:mupots_pck}
    }
    
    }

\end{floatrow} 
\end{figure*}

\begin{figure*}[t]

\begin{floatrow}

\capbtabbox[1.2\linewidth]{%

\setlength{\tabcolsep}{1.1pt}
\resizebox{1.02\linewidth}{!}{
    \caption{Joint wise analysis of 3DPCK$_{rel}$ on MuPoTS-3D (higher is better). Underlined values indicate that our unpaired learning (\textit{Ours-Us}) performs better on that joint
    } 
    
    \begin{tabular}{l|c|c|c|c|c|c|c|c|c}
        \hline
        Methods & Hd. & Nck. & Sho. &  Elb. & Wri. & Hip & Kn. & Ank. & Avg \\ 
        \hline
        Rogez\cite{rogez2017lcr} & \underline{49.4}&\underline{67.4}&\underline{57.1}&\underline{51.4}&\underline{41.3}&\underline{84.6}&\underline{56.3}&\underline{36.3}&\underline{53.8} \\
        Mehta\cite{mehta2018single} & 62.1&81.2&77.9&\underline{57.7}&47.2&\textbf{97.3}&\underline{66.3}&47.6&66.0 \\
        \rowcolor{gray!14} 
        \textit{Ours-Us} &52.9&79.0&72.2&57.9&45.3&89.9&66.9&45.1&63.3\\
        \rowcolor{gray!14}
        \textit{Ours-Ws}& 59.9&82.4&78.0&60.6&42.3&91.5&67.2&45.5&65.2\\
        \rowcolor{gray!14}
        \textit{Ours-Fs}& \textbf{63.4}&\textbf{85.5}&\textbf{84.2}&\textbf{70.4}&\textbf{56.8}&95.0&\textbf{78.2}&\textbf{59.0}&\textbf{74.0}\\
        \hline
        
    \end{tabular}
    \label{tab:jointwise}}}
 
\capbtabbox[0.83\linewidth]{%
    \setlength{\tabcolsep}{1.2pt}
\resizebox{0.99\linewidth}{!}{   
   \caption{We report Camera Centric absolute 3DPCK$_{abs}$ metric on MuPoTS-3D. B/U means Bottom-up. \textit{fps} is runtime frames/second.
   }
    \centering
    \begin{tabular}{l|c|c|c}
        \hline
        Methods & B/U &3DPCK$_{abs}$ ($\uparrow$) & \textit{fps} ($\uparrow$) \\
        \hline
        Moon* \cite{moon2019camera} & \xmark & 9.6 & 7.3 \\
        Moon \cite{moon2019camera} & \xmark & \textbf{31.5} & 7.3 \\
        \rowcolor{gray!14}
        \textit{Ours-Us} & \cmark &  23.6 & \textbf{21.2} \\
        \rowcolor{gray!14}
        \textit{Ours-Ws} & \cmark& 24.3 & \textbf{21.2}  \\
        \rowcolor{gray!14}
        \textit{Ours-Fs} & \cmark& 28.1  & \textbf{21.2} \\
        \hline
    \end{tabular}
    \label{tab:mupots_absolute}}}
{%
}
\end{floatrow} 
\end{figure*}

\subsection{Datasets and Quantitative Evaluation}
\label{sec:dataset_quant_subsection}

\begin{figure*}[t]

\begin{floatrow}
\capbtabbox{%
    \setlength{\tabcolsep}{2.5pt}
    
    \resizebox{\textwidth}{!}{
    \caption{Comparison of Absolute MPJPE (lower is better) on Human 3.6M evaluated on S9 and S11. The table is split into three parts: single-person 3D pose estimation approaches (No. 1 to 6), multi-person 3D pose estimation \textit{top-down} approaches (No. 7 to 10), multi-person 3D pose estimation \textit{bottom-up} approaches (No. 11 and 12). Our approach performs better than previous bottom-up multi-person pose estimation methods.}
    
    \begin{tabular}{l|l|ccccccccccccccc|c}
         \hline
         No.&Methods&Dir.&Dis.&Eat&Gre.&Phon.&Pose&Pur.&Sit&SitD.&Smo.&Phot.&Wait&Walk&WaD.&WaP.&Avg  \\\hline\hline
        \multicolumn{17}{c}{\textit{\textbf{Single-person approaches}}} \\
         \hline
       
        1.&Martinez \cite{martinez2017simple} &51.8&56.2&58.1&59.0&69.5&55.2&58.1&74.0&94.6&62.3&78.4&59.1&65.1&49.5&52.4&62.9
        \\
        2.&Zhou \cite{zhou2017towards} &54.8&60.7&58.2&71.4&62.0&53.8&55.6&75.2&111.6&64.1&65.5&66.0&51.4&63.2&55.3&64.9
        \\
        3.&Sun \cite{sun2017compositional} &52.8&54.8&54.2&54.3&61.8&53.1&53.6&71.7&86.7&61.5&67.2&53.4&47.1&61.6&53.4&59.1
        \\
        4.&Dabral \cite{dabral2018learning} & 44.8&50.4&44.7&49.0&52.9&43.5&45.5&63.1&87.3&51.7&61.4&48.5&37.6&52.2&41.9&52.1
        \\
        5.&Hossain \cite{rayat2018exploiting} &44.2&46.7&52.3&49.3&59.9&47.5&46.2&59.9&65.6&55.8&59.4&50.4&52.3&43.5&45.1&51.9
       \\
        6.&Sun \cite{sun2018integral} &47.5&47.7&49.5&50.2&51.4&43.8&46.4&58.9& 65.7&49.4&55.8&47.8&38.9&49.0&43.8&49.6 \\
        \hline
        \multicolumn{17}{c}{\textit{\textbf{Multi-person approaches}}} \\
        \hline
        
        7.&Rogez \cite{rogez2017lcr} &76.2&80.2&75.8&83.3&92.2&79.0&71.7&105.9
        &127.1&88.0&105.7&83.7&64.9&86.6&84.0&87.7\\
        8.&Rogez \cite{rogez2019lcr}&55.9& 60.0&64.5&56.3&67.4&71.8&55.1&55.3&84.8&90.7&67.9&57.5&47.8&63.3&54.6&63.5\\
        9.&Dabral \cite{Dabral2019MultiPerson3H}&52.6& 61.0&58.8&61.0&69.5&58.8&57.2&76.0&93.6&63.1&79.3&63.9&51.5&71.4&53.5&65.2\\
        10.&Moon\cite{moon2019camera}& 51.5&56.8&51.2&52.2&55.2&47.7&50.9&63.3&69.9&54.2&57.4&50.4&42.5&57.5&47.7&54.4 \\
        \hline
        11.&Mehta\cite{mehta2018single}&58.2&67.3&61.2&65.7&75.8&62.2&64.6&82.0
        &93.0&68.8&84.5&65.1&57.6&72.0&63.6&69.9\\
        \rowcolor{gray!14}
        12.& \textit{Ours-Fs}& 55.8&61.4&58.4&71.9&67.6&65.2&67.7&86.7&84.3&68.3&78.9&67.9&51.8&77.9&55.2 &67.9\\
        \hline
   \end{tabular} 
   \label{tab:human36}    
   }

    }
    {%
}
\end{floatrow}
\end{figure*}

\begin{figure*}[t]

\begin{floatrow}
\capbtabbox{%
\setlength{\tabcolsep}{1.8pt}

\resizebox{\linewidth}{!}{
    \caption{ 2D keypoint result comparison of our student model with teacher network on MuPoTS-3D. $\uparrow$ indicates that higher is better and $\downarrow$ indicates that lower is better.
    }
    \centering
    \begin{tabular}{l|c|c|c}
        \hline
        Methods & IoU ($\uparrow$) & 2D-MPJPE ($\downarrow$)  & 2D-PCK ($\uparrow$)\\
        \hline
        Teacher (Cao \cite{cao2017realtime}) &60.1&38.0&66.6\\
        \rowcolor{gray!14}
        \rowcolor{gray!14}
        $\mathcal {L}_{distl}$ (no $\mathcal{D}_{syn}$) &51.9&49.6&60.3\\
        \rowcolor{gray!14}
        \textit{Ours-Fs} &81.6&19.5&74.7\\
        \hline
    \end{tabular}
    \label{tab:better_student}}}

\capbtabbox{%
\setlength{\tabcolsep}{1.8pt}
\resizebox{\linewidth}{!}{
    \caption{Complexity analysis on MuPoTS-3D. B/U stands for bottom-up approach. $\uparrow$ indicates that higher is better and $\downarrow$ indicates that lower is better. 
    }
    \centering
    \begin{tabular}{l|c|c|c|c}
        \hline
        Methods & B/U & 3DPCK ($\uparrow$) & \textit{fps} ($\uparrow$) & Model size ($\downarrow$) \\
        \hline
        Mehta \cite{mehta2018single} & \cmark &70.8& 8.8 & $>$ 25.7M \\
        Moon \cite{moon2019camera} & \xmark & 82.5& 7.3 & 34.3M \\
        \rowcolor{gray!14}
        \textit{Ours-Fs} & \cmark & 75.8 & 21.2 & 17.1M \\
        \hline 
    \end{tabular}
    \label{tab:runtime}}
}
{%
}
\end{floatrow}
\end{figure*}

\noindent\textbf{MuCo-3DHP Training Set and MuPoTS-3D Test Set.}
Mehta et.al \cite{mehta2018single} proposed creation of training dataset by compositing images from 3D single-person dataset MPI-INF-3DHP \cite{mehta2017monocular}. MPI-INF-3DHP is created by marker-less motion capture for 8 subjects using 14 cameras. MuPoTS-3D \cite{mehta2018single} is a multi-person 3D pose test dataset that contains 20 sequences capturing upto 3 persons per frame. Each of these sequences include challenging human poses and also capture real world interactions of persons. For evaluating multi-person 3D person pose, 3DPCK$_{rel}$ (Percentage of Correct Keypoints) is widely employed \cite{rogez2017lcr,mehta2018single,moon2019camera}. In the root-relative system, a joint keypoint prediction is considered as a correct prediction if the joint is present within the range of 15cm. For evaluating absolute location of human joints in camera coordinates, \cite{moon2019camera} proposed 3DPCK$_{abs}$ in which a prediction is considered correct when the joint is within the range of 25cm. 
In Table \ref{tab:mupots_pck} we have compared the results of our method against the state-of-the-art methods. Our fully supervised approach yields state-of-the-art bottom-up performance (75.8 v/s Mehta \cite{mehta2018single} 70.8) while being faster than the top-down approaches. In Table \ref{tab:jointwise} we present joint-wise 3DPCK on MuPoTS-3D dataset. We compare against \cite{moon2019camera} on 3DPCK$_{abs}$ metric in Table \ref{tab:mupots_absolute} as it is the only work that reported on 3DPCK$_{abs}$. 

\textbf{Human 3.6M \cite{ionescu2013human3}} This dataset consists of 3.6 million video frames of single person 3D poses that have been collected in laboratory setting. 
In Table \ref{tab:human36}, we show results on Protocol 2: MPJPE calculation on after alignment of root. As shown in Table \ref{tab:human36}, our approach outperforms bottom-up multi-person works (Mehta \cite{mehta2018single} 69.9 v/s Ours 67.9) and performs on par with top-down approaches (Rogez \cite{rogez2019lcr} 63.5 and Dabral \cite{Dabral2019MultiPerson3H} 65.2).

\begin{figure}[t]
    \centering
    \includegraphics[width=\textwidth]{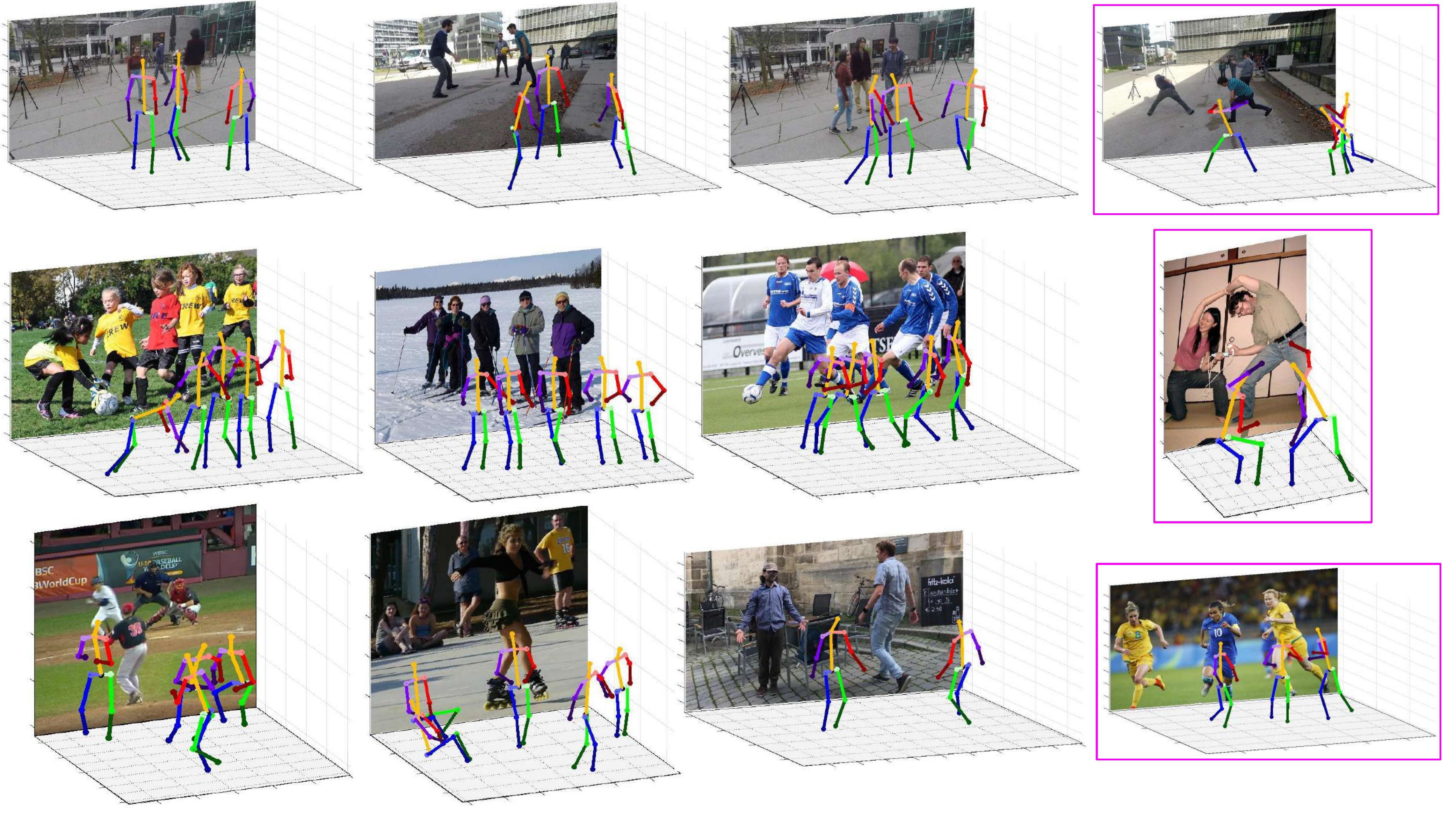}
   
    \caption{Qualitative results on MuPoTS-3D (1st row), MS-COCO (2nd row), and ``in-the-wild" images (3rd row) of our approach. Our approach is able to effectively handle inter-person occlusion and make reliable predictions for crowded images. Pink box highlights some failure cases. 1st row: presence of self-occlusion, 2nd row: rare multi-person interaction and 3rd row: joint location ambiguity. \vspace{-2mm}
    } 
    \label{fig:panel1} 

\end{figure}

\vspace{-2mm}
\section{Discussion}
\label{sec:discussion}

\noindent \textbf{Fast and accurate inference.} In Table \ref{tab:runtime}, we provide runtime complexity analysis of our model in comparison to prior works. All top-down approaches \cite{moon2019camera,rogez2017lcr,rogez2019lcr} depend on a person detector model. Hence these methods have low \textit{fps} in comparison to bottom-up approaches (See Fig.~\ref{fig:performance}). We outperform the previous bottom-up approach by a large margin in terms of 3DPCK, \textit{fps} and model size. We achieve a superior real-time computation capability because our approach effectively eliminates the keypoint grouping operation usually performed in bottom-up approaches \cite{cao2017realtime,mehta2018single}. All \textit{fps} numbers reported in Table \ref{tab:runtime} were obtained on a Nvidia RTX 2080 GPU. In Table \ref{tab:runtime}, we also show the total number of parameters of the model used during inference time.

\textbf{Is student network limited by teacher network?} In  Table \ref{tab:better_student} we report results of 2D pose estimation on both teacher model ($\hat{k}_q$) and student model ($\hat{k}_p$) by evaluating IoU, 2D-MPJPE and 2D-PCK on MuPoTS-3D dataset. We observe that a student model trained by minimizing $\mathcal{L}_{distl}$ alone performs sub-optimally in comparison to the teacher. This result is not surprising as the student model is restricted by knowledge of the teacher model. However, in our complete loss formulation (\textit{Ours-Fs}) our approach outperforms the teacher on the 2D task, validating the hypothesis that our approach can learn beyond the teacher network.

\textbf{Qualitative results.} We show qualitative results on the MS-COCO \cite{lin2014microsoft}, MuPoTS-3D and frames taken from YouTube videos and other ``in-the-wild" sources in Fig.~\ref{fig:panel1}. As seen in the Fig.~\ref{fig:panel1}, our model produces correct predictions on images with different camera viewpoints and on those images containing challenging elements such as inter-person occlusion. These qualitative results show that our model has generalized well on unseen images.  

\begin{figure}[t]
    \centering
    \includegraphics[width=0.98\textwidth]{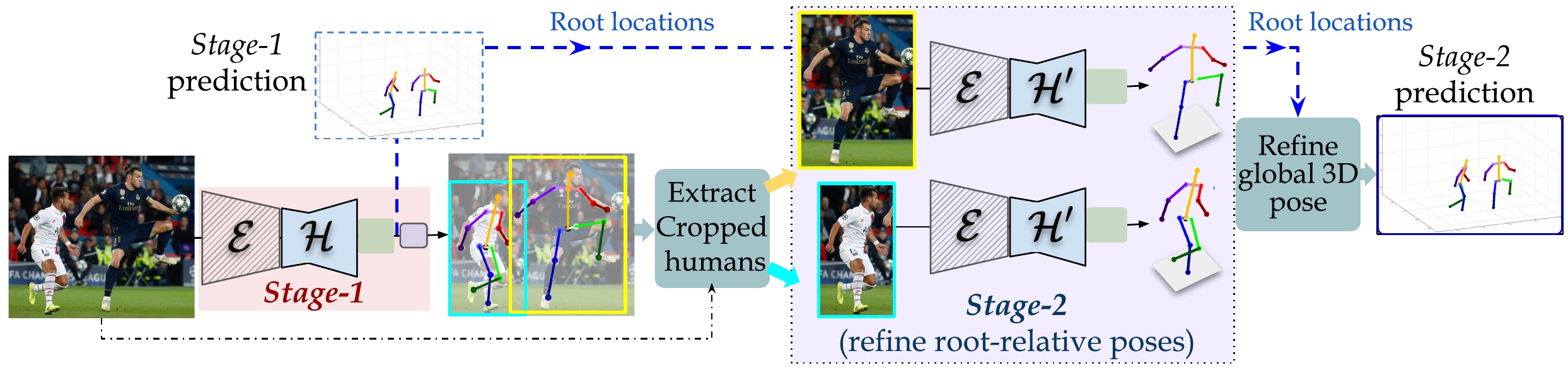} 
    \caption{A hybrid framework for two-stage refinement which treats \textit{Stage-1} output as a person detector while \textit{Stage-2} performs single-person 3D pose estimation. \vspace{-2mm}
    }
    \label{fig:stage2}
\end{figure}

\textbf{Two-stage refinement for performance-speed tradeoff.}
Top-down frameworks yield better performance as compared to the bottom-up approach while having substantial computational overhead~\cite{kocabas2018multiposenet}. To this end we realize a hybrid framework which would provide flexibility based on the requirement. For example, the current single-shot (or single-stage)  operates in a substantial computational superiority. To further improve its performance, we propose an additional pass of each detected persons through the full pipeline (Fig.~\ref{fig:stage2}). Here, we train a separate $\mathcal{H}^\prime$ for the single-person pose estimation task which is operated on the cropped image patches of single human instances obtained from the \textit{Stage-1} predictions. By training the $\mathcal{H}^\prime$ network we obtain a 3DPCK of 76.9 (v/s \textit{Ours-Fs} 75.8) with a runtime \textit{fps} of 16.6 (v/s \textit{Ours-Fs} 21.2 \textit{fps}). (See Table \ref{tab:mupots_pck} and Fig. \ref{fig:performance})

\vspace{-2mm}
\section{Conclusion}
In this paper we have introduced an unsupervised approach 
for multi-person 3D pose estimation by infusing structural constraints of human pose. Our bottom-up approach has real-time computational benefits and can estimate the pose of persons in camera-centric coordinates. Our method can benefit from future improvements on 2D pose estimation works in a plug-and-play fashion. Extending such a framework for multi-person human mesh recovery and extraction of appearance related mesh texture remains to be explored in future.

\noindent\textbf{Acknowledgement.} This project is supported by a Indo-UK Joint Project (DST/INT/UK/P-179/2017), DST, Govt. of India and a WIRIN project.

\clearpage



\section*{Supplementary material}

# Supplementary Material

## Unsupervised Cross-Modal Alignment for Multi-Person 3D Pose Estimation

The supplementary material is organized as follows:

- Section 1: Adversarial Auto-Encoder- Pose representations and training
- Section 2: Architecture and implementation details
- Section 3: Artificial poses- Sampling and analysis
- Section 4: Additional results on 3DPW dataset and 2D pose estimation
- Section 5: Limitations of the proposed framework

**Table 1.** Notation Table.

| | Symbol | Description |
|---|---|---|
| Pose Repr. | $p_g$ | 3D pose in global coordinate system |
| | $p_r$ | 3D pose in root-relative coordinate system |
| | $p_c$ | Canonical 3D pose representation |
| | $p_l$ | 3D pose in local parent relative coordinate system |
| Network | $\mathcal{E}, \mathcal{F}$ | Frozen 2D pose estimation network |
| | $\mathcal{G}$ | Encodes HM-PAF to intermediate representation |
| | $\mathcal{H}$ | Learns neural representation |
| | $\Phi, \Psi$ | Adversarial Auto-Encoder |
| | $Disc$ | Pose Discriminator used to train AAE |
| Transform-ations | FK | Forward Kinematics |
| | $\mathcal{T}_R$ | Rigid rotation operation on canonical pose |
| | $\mathcal{T}_G$ | Translation in global 3D space |
| | $\mathcal{T}_L$ | Canonical pose to local pose transformation |
| | $\mathcal{T}_K$ | Camera weak perspective projection |
| Representation (space and samples) | $\mathcal{I}$ | Image space |
| | $\mathcal{V}$ | Intermediate representation space |
| | $\mathcal{P}$ | 3D space (of multi-person pose) |
| | $\mathcal{K}$ | 2D space (of multi-person pose) |
| | $m_{syn}$ | Synthetic HM-PAF representation for 2D pose |
| | $r_x, r_y$ | Root (pelvis joint) location |
| | $s, \tilde{s}$ | Neural representation |
| | $\hat{k}_p, \hat{k}_q$ | Student and Teacher 2D pose predictions respectively |
| | $P, \hat{P}$ | Multi-person 3D pose GT and prediction |
| | $v, \tilde{v}$ | A sample in $\mathcal{V}$ space |
| Others | DoF | Degrees of Freedom |
| | $\mathcal{D}_{syn}$ | Synthetic Dataset |
| | $\theta, \gamma$ | Angle parameters in spherical coordinate system |

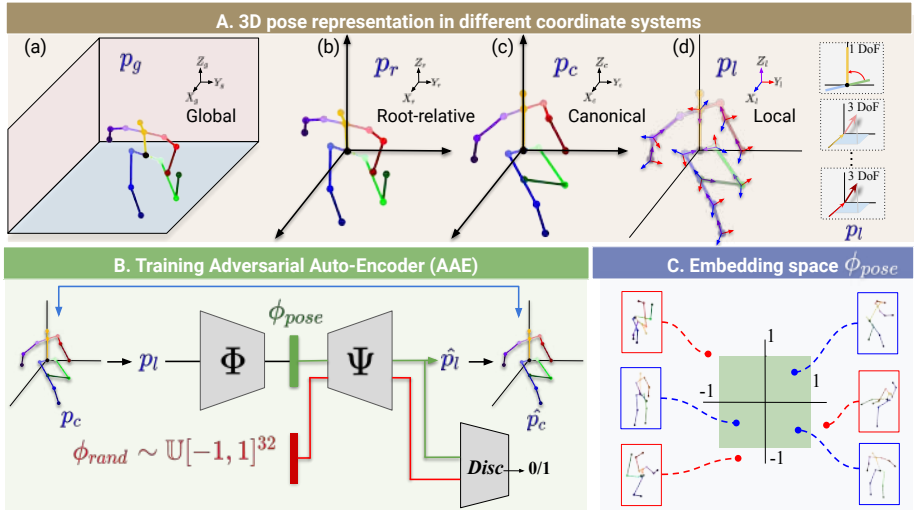

**Fig. 1. A.** 3D pose representation in 4 different coordinate systems- (a) Global, (b) Root-relative, (c) Canonical and (d) Local. On the right, DoFs are shown for certain joints. Right-hip joint has only one DoF in local coordinate system. **B.** Training framework for AAE. **C.** The AAE trained with single-person pose datasets decodes a plausible pose when sampled in $\mathbb{U}[-1, 1]^{32}$. blue box: plausible pose, red box: implausible pose

## 1  Adversarial Auto-Encoder (AAE)

We train an AAE to learn single-person pose embedding. The proposed framework for training the AAE using encoder $\Phi$, decoder $\Psi$ and adversarial discriminator ***Disc*** is shown in Fig. 1B. The main motivation behind learning the single-person pose embedding is to disentangle enforcement of structural plausibility constraints for 3D human pose in the subsequent final task of multi-person pose estimation. This parameterization of 3D pose embedding not only guarantees generation of anthropomorphically plausible pose, but also follows the structural constraints [1] such as joint angle limits, limb interpretation restrictions, etc.

**a) View-invariant Canonical 3D Pose Representation.** Let $p_g$ be a 3D pose in the global coordinate system, as shown in Fig. 1A(a). The root-relative 3D pose $p_r$ (origin of coordinate system is located at root joint) as shown in Fig. 1A(b) is obtained after subtracting human pelvis location (*a.k.a* root) from $p_g$. Then, the rigid transformation on $p_r$, disentangles the root-relative pose into view invariant canonical pose $p_c$. Let us consider a plane passing through the neck, left-hip and right-hip joints. Let $\hat{n}$ be a normal to this plane. In the canonical coordinate system, which is defined by axes $X_c$, $Y_c$ and $Z_c$ in Fig. 1A(c), the vector $\hat{n}$ is canonically aligned with +ve $X$ axis. This alignment makes the canonical pose $p_c$ view-invariant. Note that, the root-relative pose $p_r$ can be recovered from $p_c$ by performing a simple rigid transformation described by the corresponding rotation matrix. The rotation matrix itself can be described with Euler angles used to rotate $p_r$ to form $p_c$.

**b) Local 3D pose representation.** Inline with [10], the forward kinematic formulation expresses each body joint with respect to its parent joint. In the local coordinate system for each joint (see Fig. 1A(d)), the kinematic 3D structure of the human skeleton can be studied by capturing the limitations of joint movements relative to the corresponding parent joint. Further, every parent-child limb is assigned a fixed bone length. For example, the bone-length of the limb connecting the left-shoulder and left-elbow is fixed for all poses. A 3D pose expressed using this kinematic formulation is termed as local pose $p_l$ and is shown in Fig. 1A(d). As $p_l$ is obtained from $p_c$, it is both view-invariant and bone-length scale invariant. The local pose coordinate system $X_l$, $Y_l$ and $Z_l$ is defined as follows: Each joint (except neck, pelvis, left-hip and right-hip) is expressed with respect to its parent joint, or in other words, the origin of the coordinate system is fixed at the parent joint. The coordinate axes are obtained by performing Gram-Schmidt orthogonalization of a vector joining parent-child and a normal $\hat{n}$ to the plane spanning neck, left-hip and right-hip joints. The transformation from canonical pose $p_c$ to local pose $p_l$ is given as $\mathcal{T}_L : p_c \to p_l$.

**c) Training AAE.** The architecture of AAE (see Fig. 1B) is based on a kinematic tree of limb-connections mentioned in [3]. The pose embedding $\phi_{pose}$ is 32 dimensional vector and obtained through *tanh* nonlinearity. We choose to train an AAE with an aim to learn pose embedding in continuous manner. This generative approach allows us to uniformly sample any random vector as $\phi \sim \mathbb{U}[-1,1]^{32}$ and predicts an anthropomorphically plausible human pose when decoded through $\Psi$. The plausible and implausible pose pattern obtained after sampling pose embedding is shown in Fig. 1C. We employ discriminator **Disc** to distinguish between real pose embedding $\phi_{real}$ and pose embedding sampled through $\phi_{rand} \sim \mathbb{U}[-1,1]^{32}$. In order to enforce learning of an one-to-one mapping in a generative adversarial setup, we add cyclic reconstruction loss on both canonical pose $p_c$ and pose embedding $\phi_{pose}$ as follows:

$$\mathcal{L}_{cyc} = \mid p_c - \widehat{p}_c \mid + \mid \phi_{pose} - \widehat{\phi}_{pose} \mid \tag{1}$$

Where, $\widehat{p}_c = FK \circ \Psi \circ \Phi \circ \mathcal{T}_L(p_c)$, $\widehat{\phi}_{pose} = \Phi \circ \Psi(\phi_{pose})$, FK: $p_l \to p_c$ and $\mathcal{T}_L : p_c \to p_l$. We train encoder $\Phi$ using $\mathcal{L}_{cyc}$ and decoder $\Psi$ using $\mathcal{L}_{cyc} + \mathcal{L}_{adv}$ inline with [4].

## 2    Architecture

In this section, we describe network architectures of $\mathcal{E}, \mathcal{F}, \mathcal{H}, \mathcal{H}', \mathcal{G}$.

*Module $\mathcal{E}$*: We use a pre-trained model of Cao *et al.* [2] as a teacher model as shown in Fig. 2. The teacher model uses VGG19 backbone, followed by separate branches of fully convolutional layers for heatmap and PAF. The *concat* operation concatenates the outputs of these branches into an output of shape $28 \times 28 \times 1024$.

*Module $\mathcal{F}$*: We use upto stage-2 of Cao *et al.* [2] as $\mathcal{F}$. As seen in Fig. 2, there are 8 convolutional layers in both HM and PAF branches. Each branch takes the input from the corresponding output branch of $\mathcal{E}$ in the distillation pathway and output of $\mathcal{G}$ in the auto-encoding pathway (Fig. 4 of the main paper).

4   Kundu et al.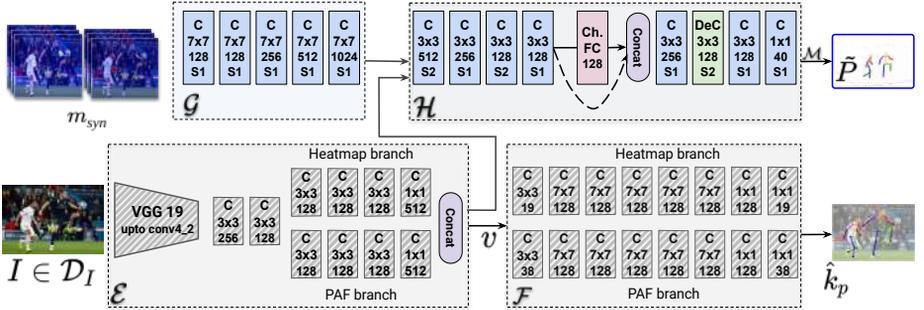

**Fig. 2.** 'C' stands for Convolutional layer. 'Ch. FC' stands for Channel-wise Fully Connected layer [7]. 'DeC' stands for Deconvolutional layer. Dashed connection indicates skip-connection. Both $\mathcal{E}$ and $\mathcal{F}$ are frozen while training $\mathcal{G}$ and $\mathcal{H}$. $S$ indicates stride.

*Module* $\mathcal{G}$: It consists of five $7 \times 7$ convolutional layers as shown in Fig. 2. The input $m_{syn}$ is of $28 \times 28 \times 43$ dimension where 15 channels correspond to each of the 15 joints and 28 channels correspond to PAF representation for all limbs.

*Module* $\mathcal{H}$ *and* $\mathcal{H}'$: Both $\mathcal{H}$ and $\mathcal{H}'$ network modules share the same architecture. These modules take an embedding $v$ as an input and predict a tensor of shape $14 \times 14 \times 39$. Further, these modules have a Channel-wise Fully Connected layer (Ch-FC) (similar to [7]) where the layer connects all nodes of a given input channel to all nodes of corresponding output channel. In our architecture, this layer takes $7 \times 7 \times 128$ as input tensor shape and outputs tensor of the same shape. Since each of the 128 channels has a spatial dimension of $7 \times 7$, the Ch-FC layer consists of 128 fully connected layers with 49 input nodes and 49 output nodes in each layer. The final layer of $\mathcal{H}$ uses an activation of *tanh* which ensures that the output space of $\mathcal{H}$ results in plausible 3D pose prediction (via $\Psi$). All other layers in the module $\mathcal{H}$ use Leaky ReLU activation.

### 2.1   Differentiable transformation operations in $\mathcal{M}$

Module $\mathcal{M}$ consists of frozen 3D pose embedding decoder $\Psi$, forward kinematics operation (FK), pose 3D rigid transformation $\mathcal{T}_R$ and 3D scene composition by translating multiple root-relative 3D poses $\mathcal{T}_G$.

**a) Forward kinematics (FK)** $p_l \to p_c$. Using forward kinematics, the local pose predicted by $\Psi$, is converted into view-invariant canonical 3D pose [1].

**b) Rigid rotation transformation** $\mathcal{T}_R : p_c \to p_r$. Module $\mathcal{H}$ predicts sine and cosine angle components for 3 angle parameters (Euler angles, denoted as $c$) required to perform rigid rotation. Using the Euler angles, the canonical pose $p_c$ is transformed to the root-relative pose $p_r$ as described in Section 1.

**c) Global scene composition** $\mathcal{T}_G : p_r \to p_g$. Using the predicted 2D root-keypoints $r_x, r_y$ and the depth $d$, the net translation of the pose is computed as a function of $(r_x, r_y, d)$. This translation is performed on 3D pose of each person as inferred in the neural representation (*i.e.* where a root-joint can be inferred).

## 2.2   Other implementation details

We develop a differentiable camera module with fixed configuration (focal lengths and center of camera are fixed based on input image size) for projecting the 3D scene. The unpaired 3D poses are normalized for keeping the bone length ratio fixed. As discussed previously, this dataset is used for training the pose decoder $\Psi$ and also used for creating multi-person 3D skeleton scenes $\mathcal{D}_{syn}$.

We first pretrain $\mathcal{H}$ using $\mathcal{L}_{distl}$ for about 15k iterations before imposing all losses. Our **phase-1** of training requires 450k iterations to converge. After training for 450k iterations, **phase-2** of our training is started. As discussed in main paper, in **phase-2** of our training, we impose only $\mathcal{L}_{ss}$ and $\mathcal{L}_{recon}$ while keeping $\mathcal{G}$ frozen.

## 3   *Artificial-pose-sampling*

Artificial poses are created by sampling from joint-angle ranges specified by a biomechanic expert. These joint-angle limits are described in the local parent-relative system on the canonical pose representation (see Fig. 1). Therefore, the poses that are sampled from these angle limits provide us with diverse canonical poses. As described in Section 1, these poses can be used to train the AAE and to create the $\mathcal{D}_{syn}$, in a completely unsupervised setting where a 3D human pose dataset is inaccessible. In this section, we describe the sampling procedure and provide an analysis of the reliability of the *Artificial-pose-sampling*.

### 3.1   Sampling Procedure

We use the joint-angle limits defined per joint in the local coordinate system and visualize the limits in Fig. 3A. As shown in Fig. 3A, every joint can be completely described in a spherical coordinate system using two angle limits (azimuth and elevation). We represent the angles as a range in azimuth $[\theta_1, \theta_2]$ where $-180° < \theta \leq 180°$ and elevation $[\gamma_1, \gamma_2]$ where $0° \leq \gamma \leq 180°$. As described in the Section 1, certain joints, such as the right hip joint has only 1DoF while some joints such as the neck joint has 0DoF. Note that 3D keypoint locations of the left hip and the left shoulder joints can be inferred in canonical pose directly without sampling, because the pelvis joint and neck joint are the mid-points of the hip joints and shoulder joints respectively.

There is one limitation in describing joint angle ranges in the spherical coordinate system: angle limits for certain joints span beyond the 180° limit of $\theta$. For such joints we propose to use angle ranges that span on the opposite side (beyond 180° into negative $\theta$) of the spherical coordinate system. For example, the $\theta$ range for the right shoulder joint is 120° and spans from $\theta_1 = 120°$, but $\theta_2$ goes beyond the 180°. Therefore, we set $\theta_2$ to a value to a value that is equivalent to 240° (which is equal to -120°).

We create artificial single-person pose dataset by sampling from these joint angle limits for all joints applying bone lengths, followed by forward kinematics operation to construct a canonical pose. For obtaining a variety of root-relative poses, we apply random rotation transformation operations on canonical poses.

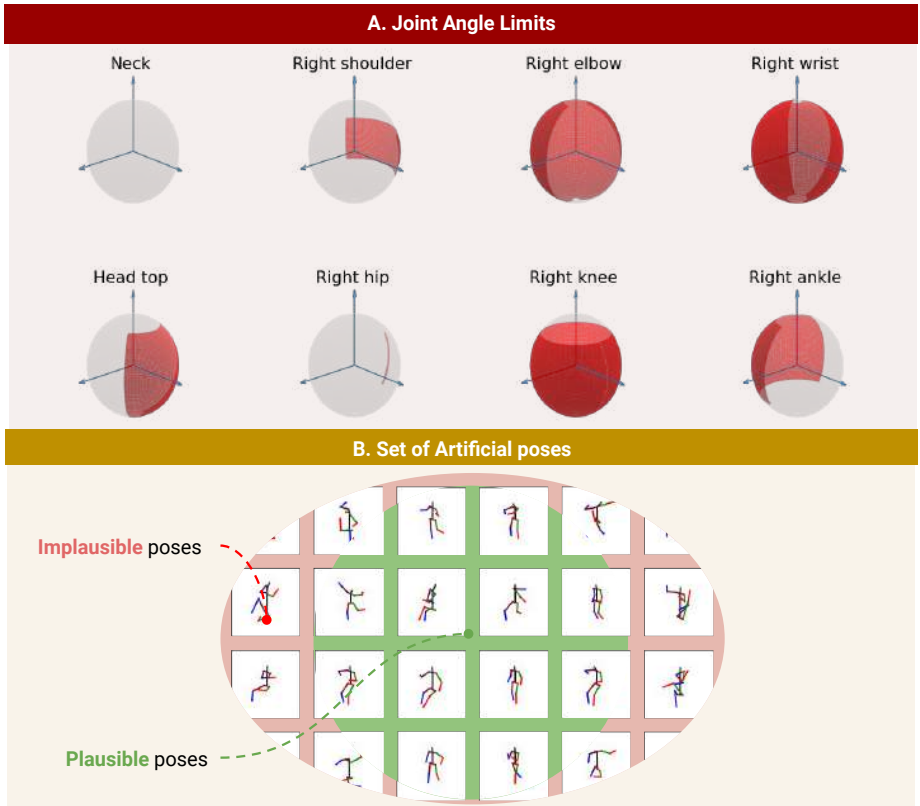

**Fig. 3.** Single-person artificial pose dataset is created by sampling uniformly from joint wise angle limits defined at local parent-relative coordinate system [1]. **A.** Since angle limits of left-body joints are symmetric to right-body joints, we present only right joints. Neck joint and right-hip joint have 0,1 DoF respectively. **B.** The artificial pose dataset subsumes all plausible poses and could contain a small fraction of implausible poses.

### 3.2   Analysis of artificial poses

Although sampled artificial poses may have certain degree of implausibility, because each joint angle is sampled independently of pose [1], we find that the artificial pose dataset subsumes all plausible poses [8,9] (see Fig. 3B). This ensures that the AAE learns rich representations in embedding space $\phi$. Our experimental analysis shown in Section 4.2 (in the main paper) confirms that having a certain degree of implausibility does not adversely affect the performance. Hence, if we are not provided an access to any unpaired 3D poses, our approach would still perform reliably by *Artificial-pose-sampling*.

## 4   Additional results

**a) Results on 3DPW dataset.** The 3D-Poses-in-the-Wild (3DPW) [5] dataset consists of challenging outdoor in-the-wild video sequences. Compared to the MuPoTS-3D dataset, the 3DPW dataset contains larger volume of video sequences

and outdoor scenes. In order to evaluate the generalizability of our model, we evaluate on the test set containing 24 sequences and show the results under the protocol *All-Test-mode*. Note that, as per the *All-Test-mode* protocol, we do not use 3DPW train set and 3DPW validation set for training our model. We use Mean Per-Joint Position Error (MPJPE) and Procrustes Mean Per-Joint Position Error as error metric (PMPJPE). The MPJPE metric is obtained as the average Euclidean distance of joints from corresponding ground-truth joint locations. In PMPJPE, the predicted pose is Procrustes aligned with the ground-truth pose before averaging the error over all joints. Therefore, PMPJPE does not consider global orientation of the predicted pose.

**Table 2.** Evaluation on 3DPW test set under the protocol *All-Test-mode*. We report MPJPE (lower is better) and PMPJPE (lower is better).

| Method | MPJPE | PMPJPE |
| --- | --- | --- |
| *Ours-Fs* | 100.7 | 77.6 |

**b) 2D keypoint prediction.** In this section, we extend the results presented in the Table 7 of the main paper. We present qualitative results in Fig. 4 to compare the 2D keypoint estimation for teacher model and student model (*Ours-Fs*) on MuPoTS-3D dataset [6]. The evaluation protocols used for 2D keypoint estimation are Intersection over Union (IoU), 2D-Mean Per-Joint Position Error (2D-MPJPE) and 2D-Percentage of correct keypoints (2D-PCK). IoU is the ratio of area of overlap between the predicted bounding box and the ground-truth bounding box to the area of union of the predicted bounding box and the ground-truth bounding box. 2D-MPJPE is average Euclidean distance between predicted 2D pose keypoints and ground-truth 2D pose keypoints. In 2D-PCK, a predicted keypoint is considered correct if it is present within a range of 25 pixels of ground-truth keypoint. All evaluations are done on keypoints that are shared by both teacher model and student model.

**c) Additional qualitative results.** We present additional qualitative results for MuPoTS-3D dataset (Fig. 6), MS-COCO 2D keypoints dataset (Fig. 7), and wild multi-person images from YouTube and other sources (Fig. 8). For MuPoTS-3D dataset, we estimate poses of all persons in the image even if ground truth annotation is absent. These results not only show that our model is able to correctly predict depth and pose of persons, but also show generalizability of our model on unseen images.

## 5    Limitations of the proposed framework

**a) Estimation of pelvis (root) location.** As discussed in Section 3.1.2 of the main paper, the neural representation of multi-person 3D pose is interpretable only in presence of a pelvis at the corresponding grid location. Therefore, in some scenarios where more than one person shares the same grid location, our model predicts only one pose for all persons in that grid. In rare cases, our model is

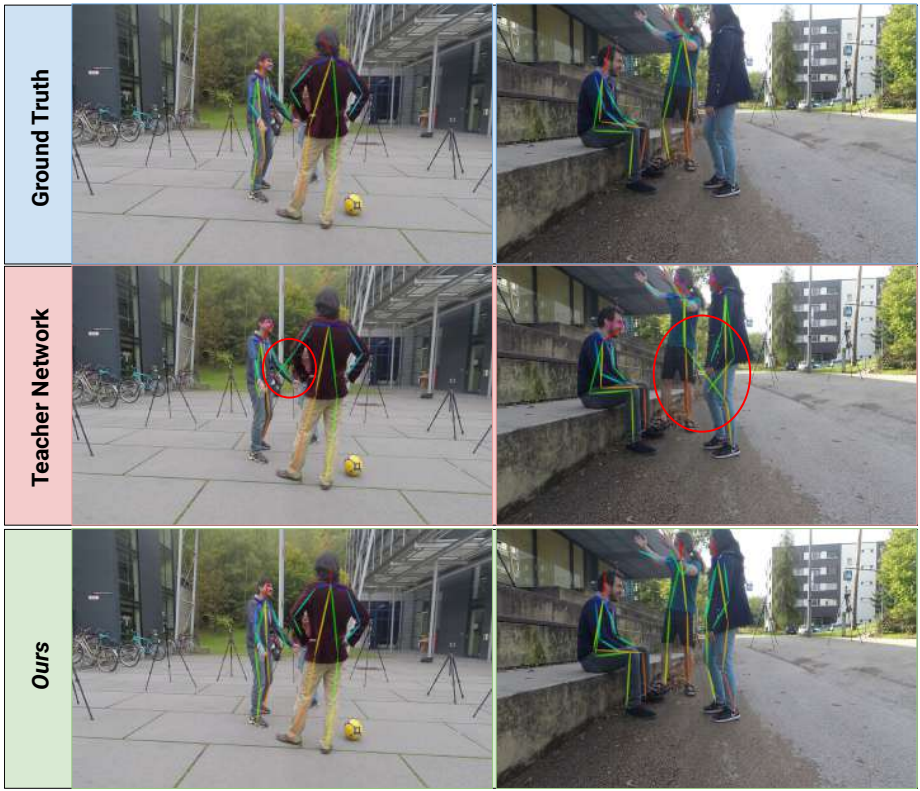

**Fig. 4.** Comparison of teacher model and student model (*Ours-Fs*) results for the task of 2D keypoint estimation on MuPoTS-3D dataset. Erroneous predictions of the teacher model are highlighted using red ovals. Teacher model either fails to predict keypoint locations or fails to assign keypoint to the correct person. As the student model estimates 2D keypoints by projecting 3D pose, it does not involve any keypoint grouping operation usually employed in bottom-up methods, such as the teacher model. These results show that the our model is able to perform better than the teacher model.

unable to predict the root joint of some persons in a given image. This limitation is shown in Fig. 5(a) and Fig. 5(b). The problem of having two pelvises in the same grid cell can be eliminated either by estimating two poses per grid-cell in the neural-representation or by increasing resolution of the output spatial map discussed in the Section 3.1 of the main paper.

**b) Rare and ambiguous poses.** Fig. 5(c) shows erroneous prediction on rarely occurring poses like acrobatic flips. The model fails to identify correct global orientation of the pose due to left-right symmetry ambiguity in lifting 2D pose to 3D pose. This limitation is also attributed to visibility of body parts. As the face of the person is not visible in the image of the Fig. 5(c), the model is not able to estimate correct body orientation. Similar example of pose ambiguity is shown in Fig. 5(d). The model predicts an ambiguous pose for the person tagged with

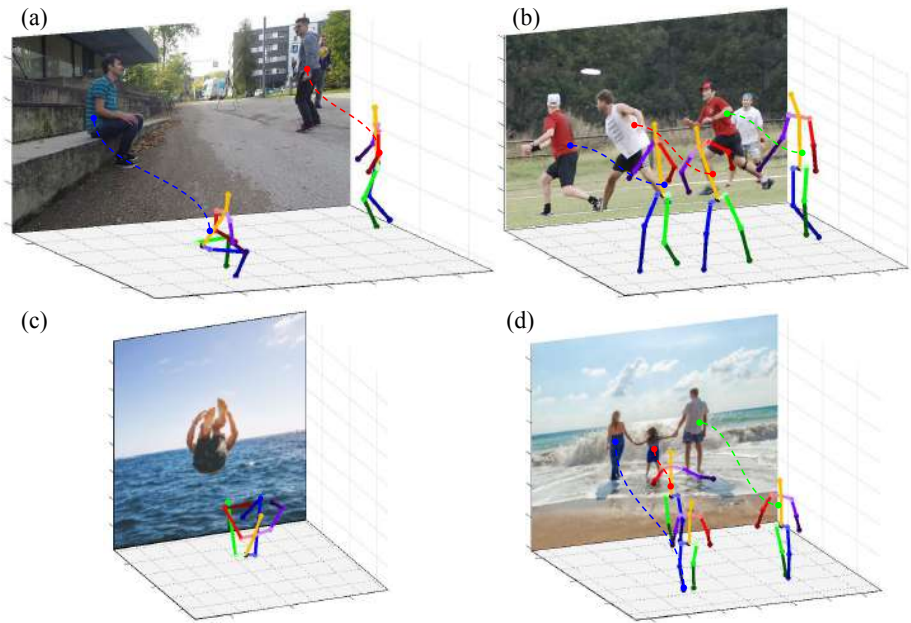

**Fig. 5.** Limitations of the proposed framework. (a) Multiple pelvises in the same grid cell, (b) Missed pelvis detection, (c) Ambiguous pose and (d) Prediction on small body-frame sized person (d) Ambiguous pose for person tagged with dashed blue line

a blue dashed line. In this case, the person's 3D pose cues in the image, such as the feet and facial orientation, are not clearly visible because of the limited spatial information owing to low-resolution of the image.

**c) Perception of depth based on bone lengths.** As the proposed model is bone-length scale-invariant, it expects all 3D poses to be of the same size. Due to this, a person with small body-frame is assumed to be located far away from the camera. This drawback is illustrated in Fig. 5(d) wherein, a person tagged with dashed red line is assumed to be of the same body-frame size as that of remaining people in the image.

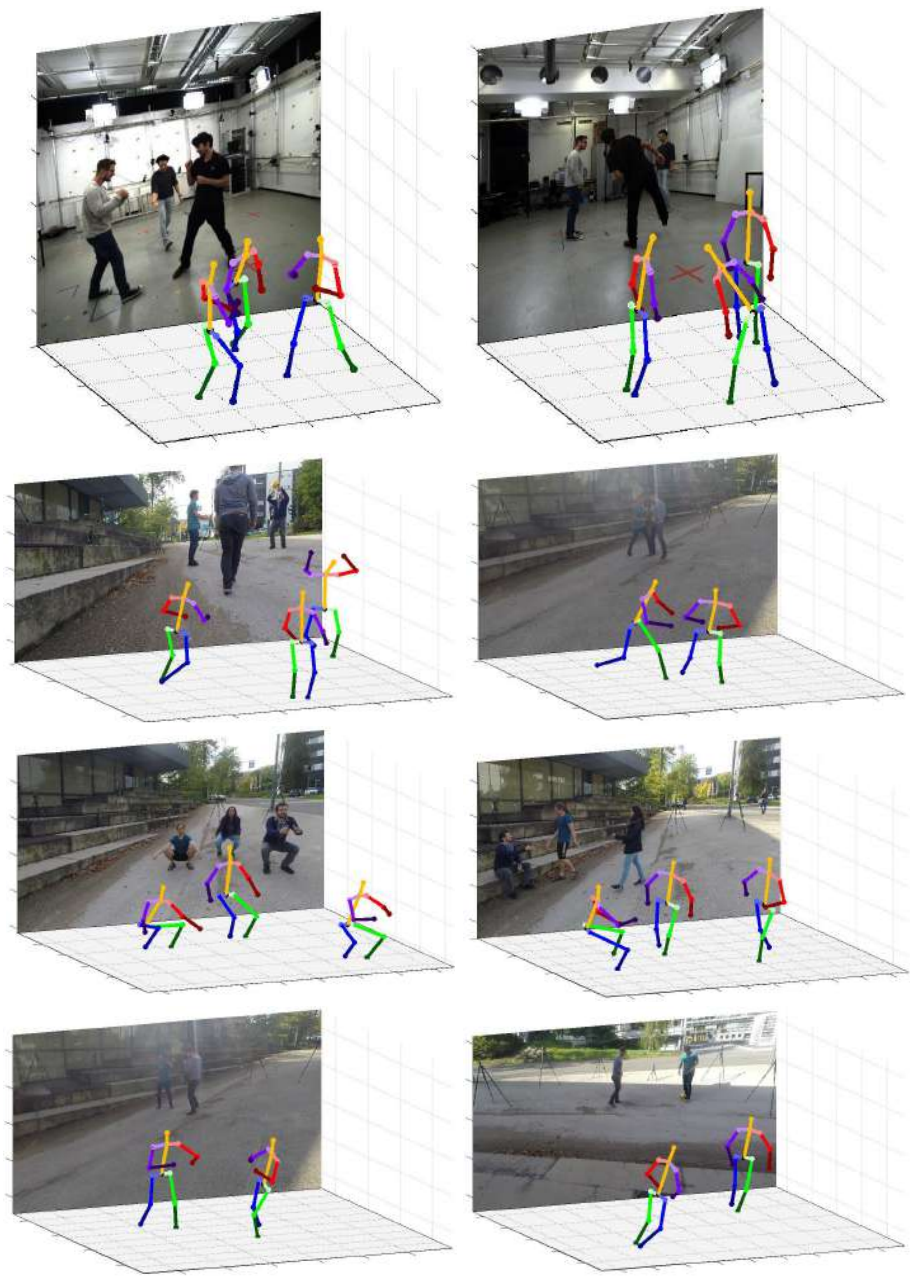

**Fig. 6.** Qualitative results on MuPoTS-3D dataset. Note that even if ground truth annotation is absent, we predict poses of all people in the image.

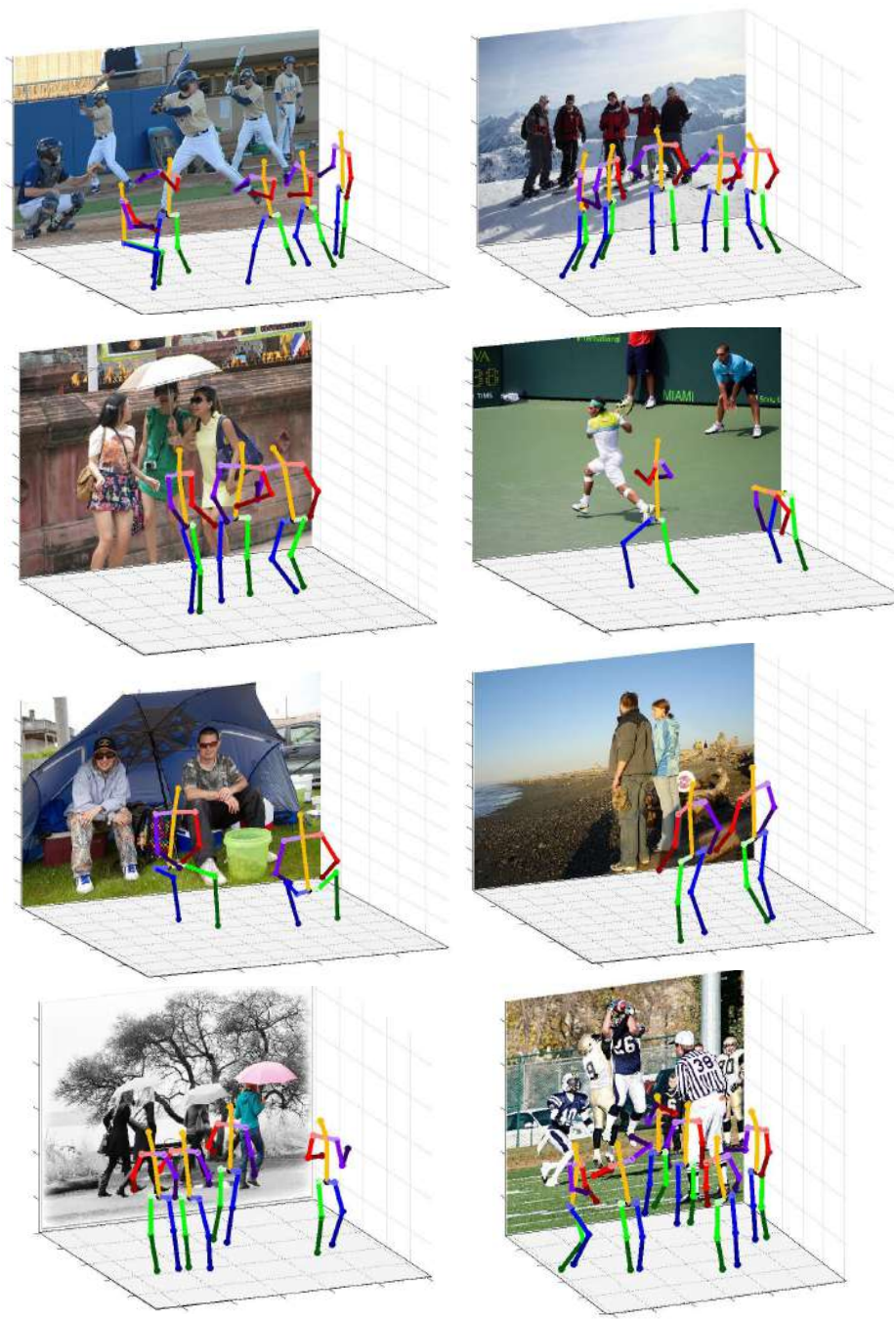

**Fig. 7.** Qualitative results on MS-COCO

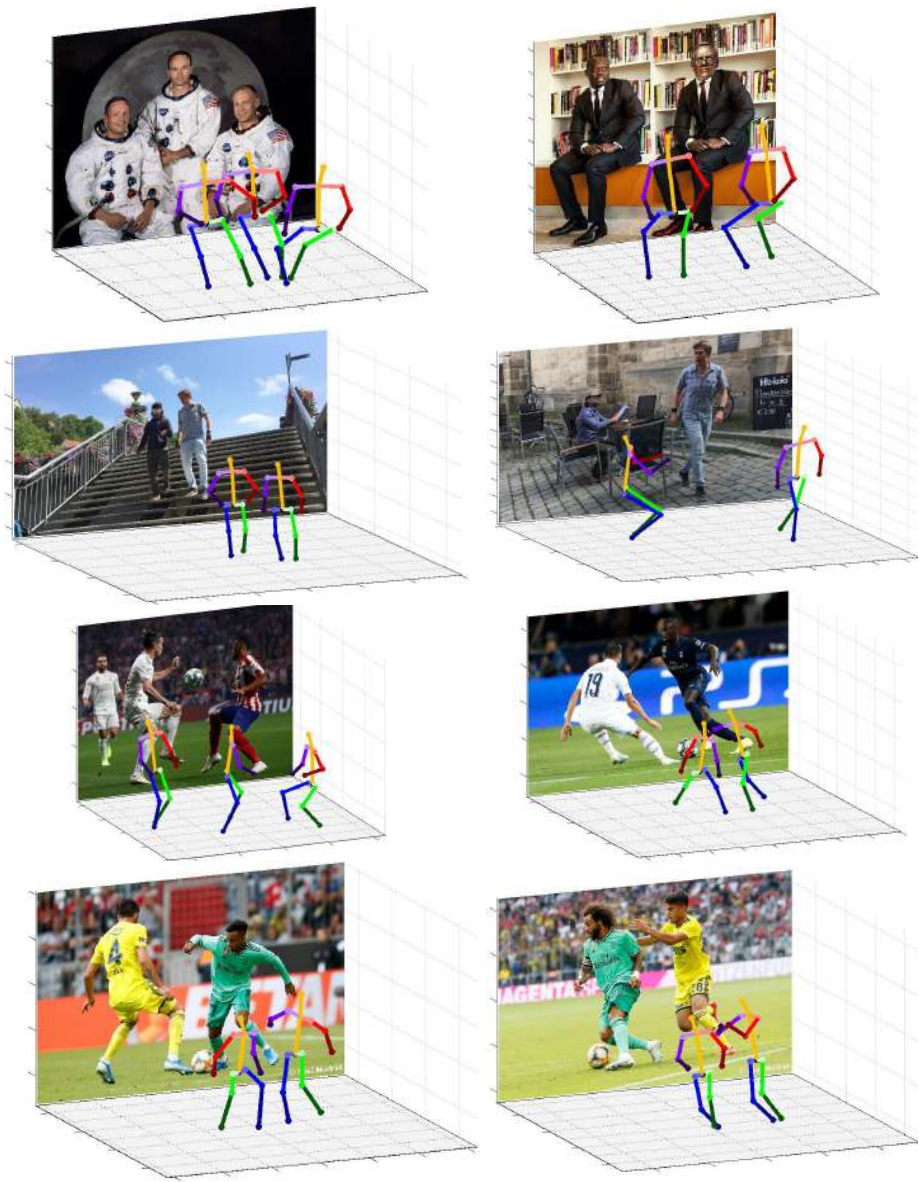

**Fig. 8.** Qualitative results on in-the-wild images

# References

1. Akhter, I., Black, M.J.: Pose-conditioned joint angle limits for 3d human pose reconstruction. In: CVPR (2015) 2, 4, 6
2. Cao, Z., Simon, T., Wei, S.E., Sheikh, Y.: Realtime multi-person 2d pose estimation using part affinity fields. In: CVPR (2017) 3
3. Du, Y., Wang, W., Wang, L.: Hierarchical recurrent neural network for skeleton based action recognition. In: CVPR (2015) 3
4. Kundu, J.N., Gor, M., Uppala, P.K., Radhakrishnan, V.B.: Unsupervised feature learning of human actions as trajectories in pose embedding manifold. In: WACV (2019) 3
5. von Marcard, T., Henschel, R., Black, M.J., Rosenhahn, B., Pons-Moll, G.: Recovering accurate 3d human pose in the wild using imus and a moving camera. In: ECCV (2018) 6
6. Mehta, D., Sotnychenko, O., Mueller, F., Xu, W., Sridhar, S., Pons-Moll, G., Theobalt, C.: Single-shot multi-person 3d pose estimation from monocular rgb. In: 3DV (2018) 7
7. Pathak, D., Krahenbuhl, P., Donahue, J., Darrell, T., Efros, A.A.: Context encoders: Feature learning by inpainting. In: CVPR (2016) 4
8. Peng, X.B., Andrychowicz, M., Zaremba, W., Abbeel, P.: Sim-to-real transfer of robotic control with dynamics randomization. In: ICRA (2018) 6
9. Tobin, J., Fong, R., Ray, A., Schneider, J., Zaremba, W., Abbeel, P.: Domain randomization for transferring deep neural networks from simulation to the real world. In: IROS (2017) 6
10. Zhou, X., Sun, X., Zhang, W., Liang, S., Wei, Y.: Deep kinematic pose regression. In: ECCVW (2016) 3




\bibliographystyle{splncs04}
\bibliography{egbib}

\begin{thebibliography}{10}
\providecommand{\url}[1]{\texttt{#1}}
\providecommand{\urlprefix}{URL }
\providecommand{\doi}[1]{https://doi.org/#1}

\bibitem{cmumocap}
{CMU} graphics lab motion capture database. available: http://mocap.cs.cmu.edu/

\bibitem{akhter2015pose}
Akhter, I., Black, M.J.: Pose-conditioned joint angle limits for {3D} human
  pose reconstruction. In: CVPR (2015)

\bibitem{cao2017realtime}
Cao, Z., Simon, T., Wei, S.E., Sheikh, Y.: Realtime multi-person 2d pose
  estimation using part affinity fields. In: CVPR (2017)

\bibitem{chen2019unsupervised_cvpr}
Chen, C.H., Tyagi, A., Agrawal, A., Drover, D., Stojanov, S., Rehg, J.M.:
  Unsupervised {3D} pose estimation with geometric self-supervision. In: CVPR
  (2019)

\bibitem{chen2018cascaded}
Chen, Y., Wang, Z., Peng, Y., Zhang, Z., Yu, G., Sun, J.: Cascaded pyramid
  network for multi-person pose estimation. In: CVPR (2018)

\bibitem{chung2018unsupervised}
Chung, Y.A., Weng, W.H., Tong, S., Glass, J.: Unsupervised cross-modal
  alignment of speech and text embedding spaces. In: NeurIPS (2018)

\bibitem{Dabral2019MultiPerson3H}
Dabral, R., Gundavarapu, N.B., Mitra, R., Sharma, A., Ramakrishnan, G., Jain,
  A.: Multi-person {3D} human pose estimation from monocular images. 3DV
  (2019)

\bibitem{dabral2018learning}
Dabral, R., Mundhada, A., Kusupati, U., Afaque, S., Sharma, A., Jain, A.:
  Learning {3D} human pose from structure and motion. In: ECCV (2018)

\bibitem{lwm}
Dhar, P., Singh, R.V., Peng, K.C., Wu, Z., Chellappa, R.: Learning without
  memorizing. In: CVPR (2019)

\bibitem{gupta2016cross}
Gupta, S., Hoffman, J., Malik, J.: Cross modal distillation for supervision
  transfer. In: CVPR (2016)

\bibitem{huang2017coarse}
Huang, S., Gong, M., Tao, D.: A coarse-fine network for keypoint localization.
  In: ICCV (2017)

\bibitem{ibrahim2016hierarchical}
Ibrahim, M.S., Muralidharan, S., Deng, Z., Vahdat, A., Mori, G.: A hierarchical
  deep temporal model for group activity recognition. In: CVPR (2016)

\bibitem{insafutdinov2016deepercut}
Insafutdinov, E., Pishchulin, L., Andres, B., Andriluka, M., Schiele, B.:
  Deepercut: A deeper, stronger, and faster multi-person pose estimation model.
  In: ECCV (2016)

\bibitem{ionescu2013human3}
Ionescu, C., Papava, D., Olaru, V., Sminchisescu, C.: Human3. 6m: Large scale
  datasets and predictive methods for {3D} human sensing in natural
  environments. TPAMI  \textbf{36}(7),  1325--1339 (2013)

\bibitem{joo2015panoptic}
Joo, H., Liu, H., Tan, L., Gui, L., Nabbe, B., Matthews, I., Kanade, T.,
  Nobuhara, S., Sheikh, Y.: Panoptic studio: A massively multiview system for
  social motion capture. In: ICCV (2015)

\bibitem{kanazawa2018end}
Kanazawa, A., Black, M.J., Jacobs, D.W., Malik, J.: End-to-end recovery of
  human shape and pose. In: CVPR (2018)

\bibitem{kingma2014adam}
Kingma, D.P., Ba, J.: Adam: A method for stochastic optimization. In: ICLR
  (2014)

\bibitem{kocabas2018multiposenet}
Kocabas, M., Karagoz, S., Akbas, E.: Multiposenet: Fast multi-person pose
  estimation using pose residual network. In: ECCV (2018)

\bibitem{nath2018object}
Kundu, J.N., Ganeshan, A., MV, R., Babu, R.V.: Object pose estimation from
  monocular image using multi-view keypoint correspondence. In: ECCVW (2018)

\bibitem{kundu2018ispa}
Kundu, J.N., Ganeshan, A., MV, R., Prakash, A., Babu, R.V.: {iSPA-Net}:
  Iterative semantic pose alignment network. In: ACM Multimedia (2018)

\bibitem{kundu2019gan}
Kundu, J.N., Gor, M., Agrawal, D., Babu, R.V.: {GAN-Tree}: An incrementally
  learned hierarchical generative framework for multi-modal data distributions.
  In: ICCV (2019)

\bibitem{kundu2019bihmp}
Kundu, J.N., Gor, M., Babu, R.V.: {BiHMP-GAN}: Bidirectional {3D} human motion
  prediction gan. In: AAAI (2019)

\bibitem{kundu2019unsupervised}
Kundu, J.N., Gor, M., Uppala, P.K., Babu, R.V.: Unsupervised feature learning
  of human actions as trajectories in pose embedding manifold. In: WACV (2019)

\bibitem{kundu2019adapt}
Kundu, J.N., Lakkakula, N., Babu, R.V.: {UM-Adapt}: Unsupervised multi-task
  adaptation using adversarial cross-task distillation. In: ICCV (2019)

\bibitem{kundu2020unsupervised}
Kundu, J.N., Patravali, J., Babu, R.V.: Unsupervised cross-dataset adaptation
  via probabilistic amodal {3D} human pose completion. In: WACV (2020)

\bibitem{kundu2020self}
Kundu, J.N., Seth, S., Jampani, V., Rakesh, M., Babu, R.V., Chakraborty, A.:
  Self-supervised {3D} human pose estimation via part guided novel image
  synthesis. In: CVPR (2020)

\bibitem{kundu2020kinematic}
Kundu, J.N., Seth, S., Rahul, M., Rakesh, M., Babu, R.V., Chakraborty, A.:
  Kinematic-structure-preserved representation for unsupervised {3D} human pose
  estimation. In: AAAI (2020)

\bibitem{lwf}
Li, Z., Hoiem, D.: Learning without forgetting. TPAMI  \textbf{40}(12),
  2935--2947 (2017)

\bibitem{lin2014microsoft}
Lin, T.Y., Maire, M., Belongie, S., Hays, J., Perona, P., Ramanan, D.,
  Doll{\'a}r, P., Zitnick, C.L.: Microsoft coco: Common objects in context. In:
  ECCV (2014)

\bibitem{long2015learning}
Long, M., Cao, Y., Wang, J., Jordan, M.: Learning transferable features with
  deep adaptation networks. In: ICML (2015)

\bibitem{dfkd}
Lopes, R.G., Fenu, S., Starner, T.: Data-free knowledge distillation for deep
  neural networks (2017)

\bibitem{luvizon20182d}
Luvizon, D.C., Picard, D., Tabia, H.: 2d/{3D} pose estimation and action
  recognition using multitask deep learning. In: CVPR (2018)

\bibitem{makhzani2015adversarial}
Makhzani, A., Shlens, J., Jaitly, N., Goodfellow, I., Frey, B.: Adversarial
  autoencoders. arXiv preprint arXiv:1511.05644  (2015)

\bibitem{martinez2017simple}
Martinez, J., Hossain, R., Romero, J., Little, J.J.: A simple yet effective
  baseline for {3D} human pose estimation. In: ICCV (2017)

\bibitem{mehta2017monocular}
Mehta, D., Rhodin, H., Casas, D., Fua, P., Sotnychenko, O., Xu, W., Theobalt,
  C.: Monocular {3D} human pose estimation in the wild using improved cnn
  supervision. In: 3DV (2017)

\bibitem{mehta2018single}
Mehta, D., Sotnychenko, O., Mueller, F., Xu, W., Sridhar, S., Pons-Moll, G.,
  Theobalt, C.: Single-shot multi-person {3D} pose estimation from monocular
  rgb. In: 3DV (2018)

\bibitem{mehta2017vnect}
Mehta, D., Sridhar, S., Sotnychenko, O., Rhodin, H., Shafiei, M., Seidel, H.P.,
  Xu, W., Casas, D., Theobalt, C.: Vnect: Real-time {3D} human pose estimation
  with a single rgb camera. ACM TOG  \textbf{36}(4),  1--14 (2017)

\bibitem{moon2019camera}
Moon, G., Chang, J.Y., Lee, K.M.: Camera distance-aware top-down approach for
  {3D} multi-person pose estimation from a single rgb image. In: ICCV (2019)

\bibitem{nayak2019zero}
Nayak, G.K., Mopuri, K.R., Shaj, V., Radhakrishnan, V.B., Chakraborty, A.:
  Zero-shot knowledge distillation in deep networks. In: ICML (2019)

\bibitem{newell2017associative}
Newell, A., Huang, Z., Deng, J.: Associative embedding: End-to-end learning for
  joint detection and grouping. In: NIPS (2017)

\bibitem{newell2016stacked}
Newell, A., Yang, K., Deng, J.: Stacked hourglass networks for human pose
  estimation. In: ECCV (2016)

\bibitem{nie2019single}
Nie, X., Feng, J., Zhang, J., Yan, S.: Single-stage multi-person pose machines.
  In: ICCV (2019)

\bibitem{pavlakos2017coarse}
Pavlakos, G., Zhou, X., Derpanis, K.G., Daniilidis, K.: Coarse-to-fine
  volumetric prediction for single-image {3D} human pose. In: CVPR (2017)

\bibitem{peng2018sim}
Peng, X.B., Andrychowicz, M., Zaremba, W., Abbeel, P.: Sim-to-real transfer of
  robotic control with dynamics randomization. In: ICRA (2018)

\bibitem{pilzer2019refine}
Pilzer, A., Lathuiliere, S., Sebe, N., Ricci, E.: Refine and distill:
  Exploiting cycle-inconsistency and knowledge distillation for unsupervised
  monocular depth estimation. In: CVPR (2019)

\bibitem{pishchulin2016deepcut}
Pishchulin, L., Insafutdinov, E., Tang, S., Andres, B., Andriluka, M., Gehler,
  P.V., Schiele, B.: Deepcut: Joint subset partition and labeling for multi
  person pose estimation. In: CVPR (2016)

\bibitem{rayat2018exploiting}
Rayat Imtiaz~Hossain, M., Little, J.J.: Exploiting temporal information for
  {3D} human pose estimation. In: ECCV (2018)

\bibitem{redmon2016you}
Redmon, J., Divvala, S., Girshick, R., Farhadi, A.: You only look once:
  Unified, real-time object detection. In: CVPR (2016)

\bibitem{rogez2017lcr}
Rogez, G., Weinzaepfel, P., Schmid, C.: Lcr-net:
  Localization-classification-regression for human pose. In: CVPR (2017)

\bibitem{rogez2019lcr}
Rogez, G., Weinzaepfel, P., Schmid, C.: Lcr-net++: Multi-person 2d and {3D}
  pose detection in natural images. TPAMI  (2019)

\bibitem{schmidt2012learning}
Schmidt, U., Roth, S.: Learning rotation-aware features: From invariant priors
  to equivariant descriptors. In: CVPR (2012)

\bibitem{spurr2018cross}
Spurr, A., Song, J., Park, S., Hilliges, O.: Cross-modal deep variational hand
  pose estimation. In: CVPR (2018)

\bibitem{sun2017compositional}
Sun, X., Shang, J., Liang, S., Wei, Y.: Compositional human pose regression.
  In: ICCV (2017)

\bibitem{sun2018integral}
Sun, X., Xiao, B., Wei, F., Liang, S., Wei, Y.: Integral human pose regression.
  In: ECCV (2018)

\bibitem{tobin2017domain}
Tobin, J., Fong, R., Ray, A., Schneider, J., Zaremba, W., Abbeel, P.: Domain
  randomization for transferring deep neural networks from simulation to the
  real world. In: IROS (2017)

\bibitem{xiao2018simple}
Xiao, B., Wu, H., Wei, Y.: Simple baselines for human pose estimation and
  tracking. In: ECCV (2018)

\bibitem{yasin2016dual}
Yasin, H., Iqbal, U., Kruger, B., Weber, A., Gall, J.: A dual-source approach
  for {3D} pose estimation from a single image. In: CVPR (2016)

\bibitem{Zheng_2017_CVPR}
Zheng, L., Zhang, H., Sun, S., Chandraker, M., Yang, Y., Tian, Q.: Person
  re-identification in the wild. In: CVPR (2017)

\bibitem{zhou2017towards}
Zhou, X., Huang, Q., Sun, X., Xue, X., Wei, Y.: Towards {3D} human pose
  estimation in the wild: a weakly-supervised approach. In: ICCV (2017)

\bibitem{zhou2016deep}
Zhou, X., Sun, X., Zhang, W., Liang, S., Wei, Y.: Deep kinematic pose
  regression. In: ECCVW (2016)

\end{thebibliography}
\end{document}